\documentclass[10pt,twocolumn,letterpaper]{article}

\usepackage{cvpr}              % To produce the CAMERA-READY version
\usepackage[accsupp]{axessibility}
\usepackage[dvipsnames,table]{xcolor}

\newcommand{\tablestyle}[2]{\setlength{\tabcolsep}{#1}\renewcommand{\arraystretch}{#2}\centering\footnotesize}

\newif\ifclean
\newcommand{\KGnote}[1]{\ifclean \else {\color{blue}{\bf KG: }#1} \fi}
\newcommand{\KG}[1]{\ifclean{#1}\else {\color{magenta}{#1}}\fi}
\newcommand{\cc}[1]{\ifclean{#1}\else {\color{green}{#1}}\fi}
\newcommand{\SXnote}[1]{\ifclean \else {\color{cyan}{\bf SX: }#1} \fi}
\newcommand{\SX}[1]{\ifclean{#1}\else {\color{purple}{#1}}\fi}

\newcommand{\KGCR}[1]{\ifclean{#1}\else {\color{blue}{#1}}\fi}
\newcommand{\ccCR}[1]{\ifclean{#1}\else {\color{green}{#1}}\fi}

\newcommand{\finalv}[1]{\ifclean{#1}\else {\color{purple}{#1}}\fi}
\newcommand{\dataset}{HowToChange} 
\newcommand{\method}{\textsc{VidOSC}}
\newcommand{\greencheck}{{\color{PineGreen}\ding{51}}} 
\newcommand{\redcross}{{\color{red}\ding{55}}}  

\newcommand{\custompar}[1]{
  \par
  \vspace{2pt}
  \noindent\textbf{#1}
}

\definecolor{LighterGray}{gray}{0.93}
\newcommand{\thickhline}{\Xhline{2\arrayrulewidth}}

\cleantrue  % uncomment this line for the clean version

\definecolor{cvprblue}{rgb}{0.21,0.49,0.74}
\usepackage[pagebackref,breaklinks,colorlinks,citecolor=cvprblue]{hyperref}
\usepackage{makecell}
\usepackage{pifont}
\usepackage{multirow}
\usepackage{arydshln}
\usepackage{booktabs}
\usepackage{float}

\def\paperID{5221} % *** Enter the Paper ID here
\def\confName{CVPR}
\def\confYear{2024}

\title{Learning Object State Changes in Videos: An Open-World Perspective}

\author{
Zihui Xue\textsuperscript{1,2}\qquad Kumar Ashutosh\textsuperscript{1,2} \qquad Kristen Grauman\textsuperscript{1,2} \\
\textsuperscript{1}The University of Texas at Austin \qquad
\textsuperscript{2}FAIR, Meta\\
}

\begin{document}
\maketitle
\begin{abstract}
   Object State Changes (OSCs) are pivotal for video understanding. \cc{While humans can effortlessly generalize OSC understanding from familiar to unknown objects,} current approaches are confined to a closed vocabulary. Addressing this gap, we introduce a novel open-world formulation for the video OSC problem. The goal is to temporally localize the three stages of an OSC---the object's initial state, its transitioning state, and its end state---whether or not the object has been observed during training. Towards this end, we develop V\scalebox{0.8}{ID}OSC, a holistic learning approach that: (1) leverages text and vision-language models for supervisory signals \cc{to obviate manually labeling OSC training data}, and (2) abstracts fine-grained shared state representations from objects to enhance generalization. Furthermore, we present \dataset, the first open-world benchmark for video OSC localization, which offers an order of magnitude increase in the label space and annotation volume compared to the best existing benchmark. Experimental results demonstrate the efficacy of our approach, in both traditional closed-world and open-world scenarios.\footnote{Project webpage: \url{https://vision.cs.utexas.edu/projects/VidOSC/}.}
\end{abstract}

\section{Introduction}
\label{sec:intro}

%\KGnote{overall intro is good but it is long.  can we tighten it up, and give more clear tie from the challenges presented by open-world OSC and what our proposed solution does.}

%\SXnote{discussion: whether we want to keep the object-centric in the motivation (see some clarification in L123-127, L257-264.}

%\KGnote{if we are emphasizing new compositions / open world the most, this can appear in title?}

% The proliferation of online instructional videos presents a rich source for large-scale video understanding, showcasing a myriad of human daily-life activities that involve a variety of objects undergoing different transitions. For example, a cooking video may capture the entire journey of a pineapple being peeled, chopped into chunks and finally sliced, and a woodworking video document the transition of a wooden plank being measured, cut and sanded.   It is in this rich visual tapestry that we identify a compelling yet under-explored facet: Object State Changes (OSCs). An OSC~\cite{} refers to a visually detectable transformation in which an object experiences a change, either temporarily or permanently, in a way that is not easily reversible.  
\begin{figure}[t!]
    \centering
    \includegraphics[width=1.0\linewidth]{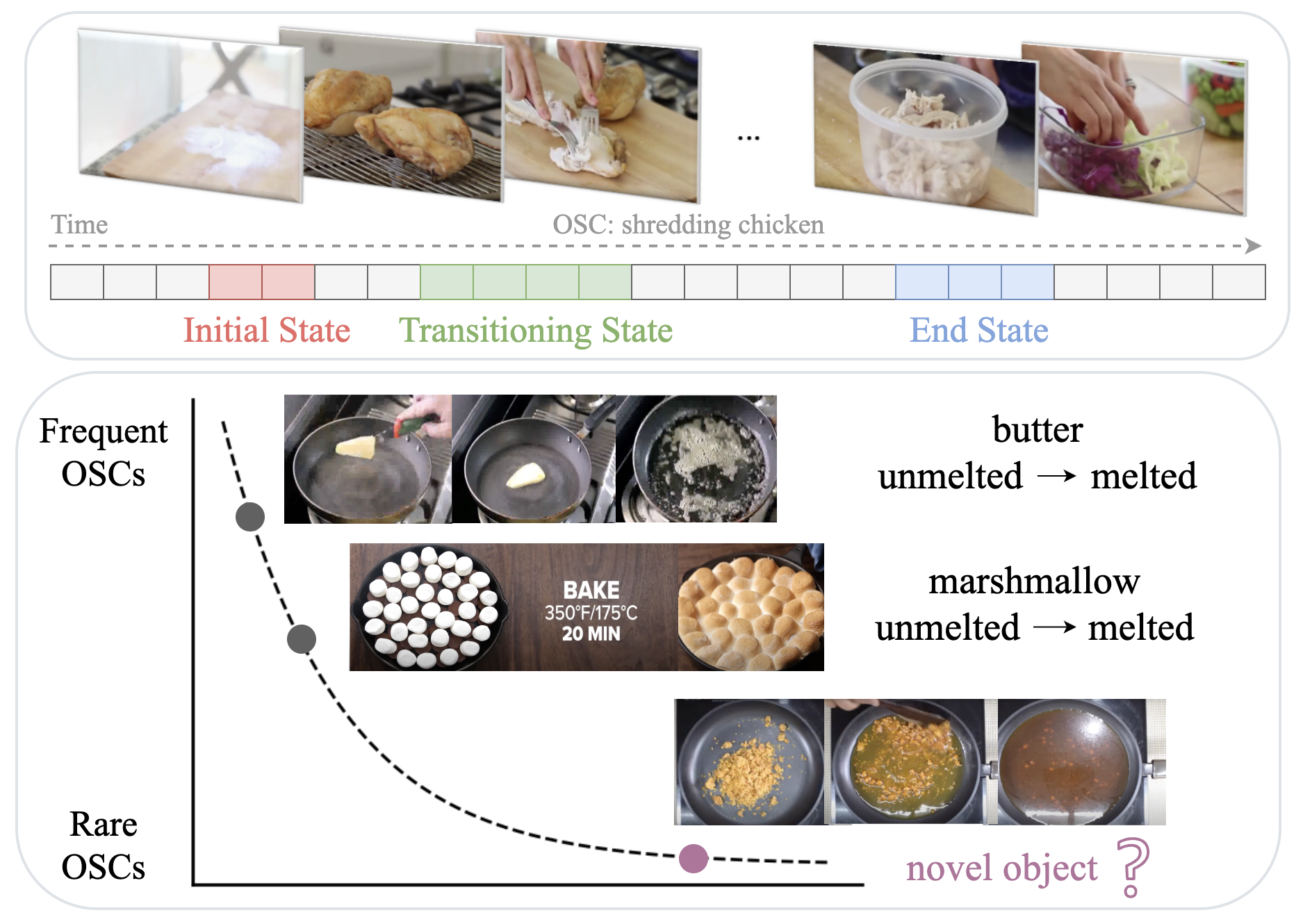}
    \caption{Top: %an illustration of the video OSC problem:
    The video OSC objective is to temporally localize %the time ranges corresponding to 
    an object's three states (i.e., initial, transitioning, end). Bottom: %In everyday scenarios, 
    \KG{OSCs naturally exhibit a long tail.} 
    Certain OSCs, such as melting butter or marshmallow, are frequently showcased in %online 
    \KG{instructional} videos while others like melting jaggery might be rarely seen. %It's crucial for a model to extrapolate its comprehension of OSCs from objects encountered during training to novel objects. Inspired by the real-world's long-tail distribution, 
    We introduce an innovative open-world formulation %for the video OSC problem 
    that requires extrapolating to novel objects never encountered during training. %\KGnote{the content of the figure is good but I wonder if it can be more beautiful, e.g., in terms of the size of the components being very easy to see.} \SXnote{updated the figure}
    }
    \label{fig:intro}
\end{figure}

In video understanding, the study of objects primarily revolves around tasks like recognition~\cite{gong2021temporal}, detection~\cite{jiao2021new}, %segmentation 
and tracking~\cite{yao2020video,ciaparrone2020deep}, %underpinned by the notion 
with the assumption that objects maintain a consistent visual appearance throughout the video. Yet, %this view is limited when considering the dynamic nature of videos, 
objects in video are often dynamic.  %Videos often 
%which often 
%feature objects undergoing state transitions, 
They can undergo transformations that change their appearance, shape, and even topology.  For example, a pineapple goes from whole to peeled to sliced, 
or wood is \KG{carved} into a new shape.  

Object State Changes (OSCs)~\cite{fathi2013modeling,mit_states,alayrac2017joint,grauman2022ego4d,souvcek2022look,souvcek2022multi,tokmakov2023breaking,saini2023chop,yu2023video,wu2023localizing}
%, as visually perceptible transformations that signify an object's change in its state, 
 add a critical dimension to video understanding. On the one hand, they provide insights into human actions---observing a piece of metal being shaped into a hook, for instance, implies the action of bending; observing an egg shell go from whole to broken implies the action of cracking.   
On the other hand, OSCs are %valuable
essential for assessing goal completion, in a way that is invariant to the specific procedure used.
%offering a view complementary to human-centric video tasks such as action recognition~\cite{jhuang2013towards,herath2017going}. This object-centric perspective facilities the learning of features that link an object's initial and end conditions during a state transition, invariant of the human action. 
For instance, the readiness of cake batter signifies the completion of a mixing task, regardless of whether it was stirred by hand or a mechanical mixer. 
%All these underscore the instrumental role of identifying OSCs in real-world applications, 
These core functionalities are instrumental for various real-world applications~\cite{heidarivincheh2017detecting,heidarivincheh2018action,epstein2020oops,epstein2021learning,chang2020procedure,bi2021procedure,zhao2022p3iv}, ranging from AR/VR assistants~\cite{plizzari2023outlook} that guide users through complex tasks by monitoring object states, to robotic manipulation~\cite{tremblay2018deep,deng2020self} where understanding the state of objects is critical for task planning and failure recovery. 

However, the development of video OSC is still at a primitive stage. Existing approaches~\cite{alayrac2017joint,souvcek2022look,souvcek2022multi,fathi2013modeling,aboubakr2019recognizing,saini2022recognizing,wang2016actions}
%\KGnote{I'd include those refs targeting actions too~\cite{fathi2013modeling,aboubakr2019recognizing,saini2022recognizing}, since they have closed world also, e.g. the action is paired with only one object, like make sandwich by spreading peanut butter.  it's better if we can critique a larger set of refs at once.} %\KGnote{ego4d here too?} \SXnote{Ego4d is different, it is OSC-agnostic localization and does not have a OSC vocabulary.} work with a 
assume a %constrained OSC 
\KGCR{\textbf{\KG{closed} vocabulary}}, where each state transformation is associated with a limited set of objects---often \KG{just 1 or 2}. %, and} no more than 5. 
For example, the concept of ``melting" might be limited to familiar items like butter and marshmallow. Consequently, the learned models are only capable of identifying state changes for objects observed during training and stumble when presented with novel objects.  % reflection of a closed-world setting. 
In contrast, in real-world situations %display a myriad of complexities. A
a single state transition like ``melting" can be linked with a plethora of objects---some ubiquitous and others rare. Importantly, %Yet
there is an intrinsic connection that threads these OSC processes together. As humans, even if we have never encountered an unusual substance (e.g., \SX{jaggery}) % a rare mineral) 
before, we can still understand that it is experiencing a ``melting'' process based on its visual transformation cues (see Fig.~\ref{fig:intro}, bottom). 

%, without any prior knowledge about the object itself. 
% This raises the need of exploring video OSCs in an open-world context.

%\begin{itemize}
%    \item \KGnote{terminology: consider if we want to use both the terms closed/open world, as well as ``new compositions"; make the connection explicit, or pick one way to refer to it.  And spell out the significance with an example, e.g., we may see butter melting and ice melting, but then have to recognize when we see chocolate melting. } \SXnote{Maybe use open world? seem more general than the composition term since object can be completely unseen.}
%\item    \KGnote{question: does our model necessarily need to see chocolate in some other context (e.g., chocolate being chopped) to handle that example?  or it can be entirely unseen? } \SXnote{it can be entirely unseen.}
%\item \KGnote{
%let's motivate clearly how important new compositions are for OSC specifically, since related things happen to different objects...maybe give some back of the envelope numbers, e.g., for N state change verbs and M object types on average per verb, how many models we would train in naive case vs. composable case?  does this sound impressive to motivate?  and/or point to the long tail that naturally occurs - we see some objects all the time with a given SC, but there are many others we'll see less frequently and still matter.}
%\end{itemize}

In light of these limitations, % above, in this work, 
we propose a %comprehensive exploration of the video OSC problem, in an 
a novel open-world \KG{formulation of the video OSC problem.}  %context 
%and from an object-centric perspective. Towards this end, 
We formally characterize OSCs in videos 
in terms of three %carefully defined 
states that must be temporally localized: initial, transitioning, and end (see Fig.~\ref{fig:intro}, top). 
%To capture the real-world richness of OSCs, we propose a novel open-world problem formulation: The objective is to identify the time ranges corresponding to the three OSC states within a given video, as illustrated in top Fig.~\ref{fig:intro}. 
In our \KGCR{\textbf{open-world setting}}, there are \emph{known} objects and \emph{novel} objects.  
%We categorize objects involved in a state transition, such as frying, into known categories (like chicken) and novel categories (like cauliflower). 
The model is only presented with known objects during training (e.g., frying chicken, frying onions).  Then, it is evaluated on both known and novel objects (e.g., frying cauliflower). %to test its capability in the open-world setting.
\KG{In addition to this fundamental open-world generalization, we also tackle a more comprehensive notion of the transitioning state.}
Specifically, % importantly 
our transitioning states encapsulate not only action-induced modifications to the object (e.g., peeling), but also passive transformations the object undergoes \KG{without human interference} (e.g., melting) \KG{and edited-out transformations (e.g., the cake goes from raw to baked even if baked off camera).}
Though largely overlooked in existing approaches~\cite{fathi2013modeling,wang2016actions,aboubakr2019recognizing,saini2022recognizing}, \KG{these cases are both common and technically interesting, since they force a model to reason about the \emph{effects} of the core state change rather than search for evidence of a human actor carrying out the action.}

Armed with this new problem formulation, we propose a holistic video OSC learning approach, anchored by two innovative ideas.  First, 
% \SX{we explore text}
\KG{we explore text and vision-language models} \SX{(VLMs)}
for supervisory signals \KG{during training}.
% we deploy the \KG{narrated} text modality \KG{in instructional videos} as a supervisory signal. 
%\SXnote{see clarification} 
%\KGnote{tie the challenges more explicitly to our solution, e.g., why this is a good idea for handling long-tail.} \SXnote{this is not very related to open-world challenge, more of a solution on how we conduct training without ground truth labels. Clarification: prior works (souvek et al.) do not require ground truth labels for training as well, their supervision signal comes from the casual ordering assumption that an OSC starts from initial state, followed by action, then end state. Different from them, we propose to use VLMs and text to provide supervision signals.} 
% While obtaining frame-wise state annotations for vast amount of data is resource-intensive, many instructional videos are accompanied with ASR transcriptions that often provide implicit cues about ongoing OSCs. For example, a narration in a cooking video such as ``Next, I'm going to chop the tomatoes" signals an upcoming state change for the tomato. Utilizing recent developments in large language models (LLMs)~\cite{touvron2023llama,touvron2023llama2} and vision language models (VLMs)~\cite{clip,jia2021scaling,yu2022coca}, we exploit text as a pivotal supervisory signal to facilitate the large-scale training of our video OSC model. Our approach eliminates the need of exhaustive label collection for training and demands \emph{only the video---no accompanying text---}as input during test for maximum flexibility and applicability.
We pioneer the use of textual state descriptions to generate %\SXnote{see clarification} 
\KG{a long tail of} %\KGnote{useful to connect back to our theme? because we generate more pairs of object+transition from mining the text?  more unique objects pop up?} \SXnote{I'm not sure how the pseudo-labels are related to long-tail, long-tail is more about the distribution of videos used for training}
% \SXnote{still feel that long-tail does not very suit here and may be confusing} \KGnote{see text I added in 3.3}
object state pseudo-labels, by leveraging the remarkable capabilities of \SX{VLMs~\cite{clip,jia2021scaling,yu2022coca} and large language models (LLMs)~\cite{touvron2023llama,touvron2023llama2,openai2023gpt4}}. This strategy eliminates the need for exhaustive label collection for training data, facilitating large-scale model training.\footnote{Note that at test time, our model requires only the video, with no need for additional text, ensuring utmost flexibility and applicability.} 
% (2) Object-agnostic state prediction. To address the open-world challenge, we propose three modeling techniques: a shared state vocabulary, temporal modeling and object-centric features. Collectively, these designs enable our model to establish a more generalized understanding of object states and their temporal transitions, decoupling OSC state understanding from specific object instances. 
%(2)
Second, to confront the open-world challenge, we propose object-agnostic state prediction, %Our model confronts the open-world challenge by integrating 
consisting of three key techniques: a shared state vocabulary to unify the label space across objects sharing the same state transition (e.g., melting butter and melting marshmallow), temporal modeling to grasp the progression of state changes over time, and object-centric features that better represent 
the objects during an OSC. 
These designs equip our model with the ability to generalize state understanding %observed in 
\KGCR{from} known objects to novel ones.  %\cc{,based on shared characteristics that transcend specific object instances}.
%\KGnote{this is an important part but we could use more clear rationale why our insight resolves the challenge of open world, i.e., how we are capturing how the state change on certain objects can be generalized to even unseen ones because of our model.} \SXnote{revised} 
%In all, 
% We term our \textbf{O}pen-world \textbf{A}nd \textbf{T}ext-\textbf{S}upervised video \textbf{OSC} approach as \textbf{\method}. \KGnote{new name?}
\SX{We term our approach \method.}

Complementing this, we present a large-scale real-world dataset \dataset. Sourced from %online instructional videos within 
the HowTo100M collection~\cite{miech2019howto100m}, 
%\dataset\ features 409 unique object-state transition compositions spread across 41,545 videos and is further enriched with a %temporally
%\KG{manually} annotated evaluation set of 5,424 videos.   %To our understanding, 
%\dataset\ 
it sets a new benchmark in the field
\KG{of unprecedented scale and an authentic long-tail distribution,} %\KGnote{reduced a lot here since we talk about details elsewhere}
% : it not only offers an unprecedented scale—with 9.3x more OSC categories and 8.1x more annotated video instances compared to the previous largest collection~\cite{souvcek2022look}, but it also provides an authentic reflection of the real world's long-tail distribution. 
% \cc{Any given state transition, like slicing, frequently correlates with certain common objects (such as apples, cucumbers) in numerous videos, and also occasionally associates with rarer objects (such as apricots, cantaloupes) in a few videos.}
\cc{setting it apart from the constrained scope and closed-world setting of earlier benchmarks (see Table \ref{tab:dataset}).} 
Finally, experimental results demonstrate the efficacy of \method, surpassing the state-of-the-art in both traditional closed-world and novel open-world scenarios by a great margin.

%\KGnote{aside from claiming better scale, we should point out how our data really captures the long tail, vs. how existing are artificially balanced and closed world.} \KGnote{be clear we are presenting this as a validation/test set, as our training is not done with these new labels?} \SXnote{revised; yes new labels are only for evaluation.}

\section{Related Work}

\custompar{Object State Changes} %Existing approaches for modeling object states and their changes fall broadly into two main categories: image-based and video-based.
Image-based methods explore the compositional nature of objects and their attributes, including zero-shot recognition of unseen combinations%Research in this field began with treating state identification as a fine-grained classification task~\cite{farhadi2009describing,duan2012discovering}, later evolving to 
%address the zero-shot compositional challenge of objects and their states
~\cite{misra2017red,nagarajan2018attributes,mit_states,purushwalkam2019task,mancini2021open,naeem2021learning,pham2021learning,li2022siamese}, but %However, these approaches 
do not consider the temporal progression of OSCs in video, which brings new challenges. % Addressing this in the video domain requires the temporal localization of fine-grained OSC states and thus brings new challenges. 
Video-based methods %~\cite{fathi2013modeling,wang2016actions,alayrac2017joint,aboubakr2019recognizing,saini2022recognizing,souvcek2022look,souvcek2022multi} % have been developed to model state changes within a video. A few works
develop human-centric models that %propose to characterize actions through state changes, where the emphasis is to 
leverage state change cues to %enhance understanding of the manipulation action in videos
\KG{facilitate action recognition}
~\cite{fathi2013modeling,wang2016actions,aboubakr2019recognizing,saini2022recognizing},
or explore joint discovery of object states and actions
% \cc{,leveraging the temporal order of pre/post-condition states and the intervening action as weak supervision}
~\cite{alayrac2017joint,souvcek2022look,souvcek2022multi}. 
%, Alayrac et al.~\cite{alayrac2017joint} assumes a consistent temporal order in states and actions (initial state, followed by action, leading to the end state). Following this thought, recent works~\cite{souvcek2022look,souvcek2022multi} utilize the concept of temporal ordering as supervision signals to temporally localize both states and state-modifying actions. %\KGnote{what is the self-supervision aspect?} \SXnote{they use the temporal order (initial appear before action and then end state) as supervision; no ground-truth labels for training} 
Notably, %these works have been largely human-centric and
all the existing methods assume a closed world, recognizing only the OSC categories \KG{(``known" objects)} seen during training.  Our approach distinguishes itself in three crucial ways: (1) we introduce a novel open-world\footnote{\KGCR{also called ``unseen compositions" in object-attribute learning~\cite{misra2017red,materzynska2020something}.}} formulation, where the objects encountered during evaluation can be entirely unseen during training; (2) we adopt an object-centric perspective, allowing for scenarios where OSCs occur with no observable human actions in the video; and
%\KGnote{check - even though alayrac or souveck may assume this, I'm not sure the older works do.} %\SXnote{clarification: similar to the two paragraphs in intro: there are two lines of works: (1) these works~\cite{alayrac2017joint,souvcek2022look,souvcek2022multi} align with ours, they also define OSC initial and end states (but not the transitioning states, they call it action); (2) these works~\cite{fathi2013modeling,wang2016actions,aboubakr2019recognizing,saini2022recognizing} use OSCs but their task is still action recognition, so I feel that they are essentially human-centric since the final goal is to recognize human actions (they utilize OSCs as action cues)} 
(3) we propose to utilize the text modality and VLMs as supervisory signals, greatly scaling up model training and boosting performance.

Early datasets for video OSC capture a limited array of OSCs in fewer than 1,000 videos
%Early efforts
~\cite{alayrac2017joint,liu2017jointly}.  The more recent %. capture a limited array of OSCs in under a thousand videos. The recent introduction of the
ChangeIt dataset~\cite{souvcek2022look} marks an advance in dataset scale \KG{(34K videos)}, yet is still restricted to a small OSC vocabulary \KG{of 44 total object-state transitions.  It lacks the scope to adequately test unseen objects, since most state changes coincide with only 1 or 2 objects across all the videos} % with an inadequate coverage of object-state transition combinations 
(see Table \ref{tab:dataset}).  %While several
Other datasets explore different aspects of video OSCs, including object state recognition~\cite{saini2023chop}, point-of-no-return (PNR) frame localization~\cite{grauman2022ego4d}, and object segmentation~\cite{tokmakov2023breaking,yu2023video}, but they lack any temporal annotations of fine-grained OSC states needed for our task's evaluation.  \KG{Our \dataset~dataset increases the OSC vocabulary space by an order of magnitude and allows for the first time rigorous study of the open-world temporal OSC challenge.} %\SXnote{add temporal understanding after first time? since there are recent works discussing open-world OSC}

%In comparison to the largest existing dataset, ChangeIt, we expand the OSC vocabulary by 9.3x and the number of annotated instances by 8.1x. To the best of our knowledge, this represents the largest and most extensive video OSC dataset collection, featuring fine-grained temporal annotations for object states (see Table \ref{tab:dataset}).
% \KGnote{add actions = transformations paper?} \SXnote{added}
%\textbf{Methods}

% In contrast, our work adopts an object-centric perspective, allowing for scenarios where OSCs occur with no observable human actions in the video (e.g., butter passively melting in a heated skillet). Moreover, we introduce a novel open-world formulation where the objects encountered during testing are entirely unseen during training.

\custompar{Vision and Language}
\SX{LLMs~\cite{touvron2023llama,touvron2023llama2,openai2023gpt4} have revolutionized various research fields with their exceptional performance across diverse applications. Building on this momentum, the use of web-scale image-text data has emerged as a powerful paradigm in computer vision, %and led to 
\KGCR{with} powerful VLMs \KGCR{now advancing}}
% Utilizing web-scale image-text data %for learning visual representations has proven to be a 
% is a powerful paradigm in computer vision, and  VLMs~\cite{clip,jia2021scaling,yu2022coca} have advanced %the state-of-the-art in a diverse 
an array of image tasks, including zero-shot classification~\cite{clip}, detection~\cite{gu2021open}, segmentation~\cite{luddecke2022image} and visual question answering~\cite{li2022blip,li2023blip}. %Similar trends are also witnessed in the video domain. 
Similarly, joint video-text representation learning~\cite{miech2019howto100m,lin2022egocentric,zhao2023learning,ashutosh2023hiervl} has been advanced by multimodal datasets like HowTo100M~\cite{miech2019howto100m} and Ego4D~\cite{grauman2022ego4d}, which offer large-scale collections of videos paired with their corresponding narrations.
% multimodal datasets like HowTo100M~\cite{miech2019howto100m} and Ego4D~\cite{grauman2022ego4d}, which offer large-scale collections of videos paired with their corresponding %ASR transcriptions or 
% narrations, serve as a rich source for %research in joint 
% video-text representation learning~\cite{miech2019howto100m,lin2022egocentric,zhao2023learning,ashutosh2023hiervl}. 
\cc{These datasets have also facilitated %advances %in 
\KGCR{tasks} 
like step discovery~\cite{dvornik2023stepformer}, localization in procedural activities~\cite{mavroudi2023learning} and long egocentric videos~\cite{ramakrishnan2023naq}.}
%\KGnote{should we cite Ego4D+narrations line of work here also?} \SXnote{Added}
%Closely related to us, Epstein et al.~\cite{epstein2021learningdynamic} propose a 
\KGCR{The} vision-language multimodal cycle consistency loss \KGCR{proposed in~\cite{epstein2021learningdynamic} helps} discover long-term temporal dynamics in video.
In line with these developments,
we propose to leverage the text accompanying instructional videos as well as existing high-performing VLMs to provide supervisory signals that guide the training of our video OSC model \KG{and allow learning for the open world}. 

\custompar{Learning in an Open World}
The open world setting has received increasing attention, % in recent years,
predominantly within the image domain for object recognition~\cite{scheirer2012toward,bendale2015towards}, detection~\cite{dhamija2020overlooked,joseph2021towards,gu2021open,zareian2021open} and object-attribute compositionality~\cite{misra2017red,nagarajan2018attributes,purushwalkam2019task,mancini2021open,naeem2021learning,pham2021learning,li2022siamese}. In the video domain, compositionality is advanced by the Something-Else dataset~\cite{materzynska2020something}, where the training combinations of verbs and nouns do not overlap with the test set, sparking %subsequent studies on capturing 
work on the dynamics of subject-object interactions~\cite{materzynska2020something,saini2023chop,chatterjee2023opening}. More recent efforts leverage VLMs for zero-shot action recognition
%, aiming for generalization to unseen verbs
~\cite{wang2021actionclip,ju2022prompting,chatterjee2023opening}. %or nouns~\cite{chatterjee2023opening}. 
%For OSC, 
\KG{Concurrent work explores
video object segmentation with state-changing objects~\cite{yu2023video} and recognition of novel object-state compositions for food chopped in different styles~\cite{saini2023chop}.}
%recognition of novel object-state compositions~\cite{saini2023chop} an open-vocabulary state-changing object segmentation~\cite{yu2023video}, respectively. 
Despite these advances, temporal video OSC understanding in the open world remains unexplored. Our open-world formulation %diverges from existing paradigms and brings new challenges that 
requires generalizing the temporal localization of fine-grained object states from known to novel objects. 

%\KGnote{Check: are we adequately citing ``open world" papers, like for open world object recognition/detection?  a reviewer with ``open world" papers is likely to get this one, even if they don't do state change stuff. here's a recent workshop: https://www.cs.cmu.edu/~shuk/vplow.html that may have key authors/refs.} \SXnote{Updated this paragraph to be ``learning in an open world'', added open-world papers besides compositionality}
\section{Approach}

%\KGnote{We focus on cooking tasks in instructional how-to videos because they offer a wealth of objects, tools, and state changes, providing an excellent test bed for open-world OSC....}%\SXnote{discussed in Sec. \ref{sec:dataset} footnote, do we want to put this in method instead?}

%\KGnote{...Any given state transition, like slicing, frequently correlates with certain common objects (such as apples, cucumbers) in numerous videos, and also occasionally associates with rarer objects (such as apricots, cantaloupes) in a few videos. This stands in stark contrast to the constrained scope and closed-world setting of earlier benchmarks.} \SXnote{similar contents in Sec.\ref{sec:problem_formulation}}% (see Table \ref{tab:dataset})....

\begin{figure*}[t!]
    \centering
    \includegraphics[width=0.95\linewidth]{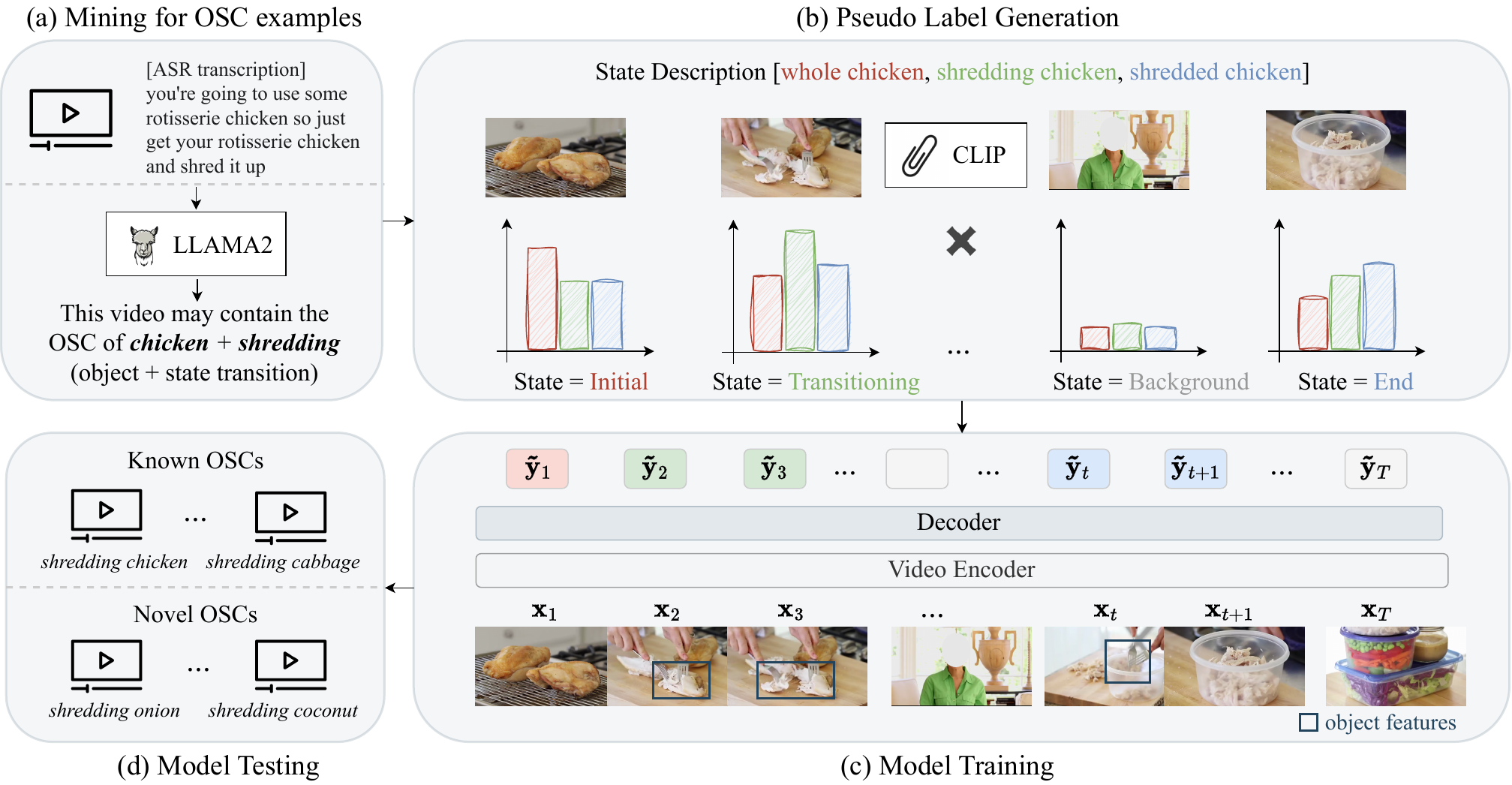}
    \vspace*{0mm}
    \caption{Our proposed \method\ framework: \SX{(a) Mining for OSC examples  (Sec. \ref{sec:pseudo_label}): We leverage ASR transcriptions paired with videos and the capabilities of LLM to automatically mine OSC examples.}
    % We automate the process to create the \dataset\ dataset, leveraging ASR transcriptions paired with videos and the capabilities of LLM. 
    (b) Pseudo Label Generation (Sec. \ref{sec:pseudo_label}): We \SX{utilize textual state descriptions and a VLM} for supervisory signals during training; (c) Model Training (Sec. \ref{sec:model_design}): We develop a video model for object-agnostic state prediction. (d) Model Testing (Sec. \ref{sec:problem_formulation}): We propose an open-world formulation, evaluating on both known and novel OSCs. Notably, while we employ the text modality to guide model training, our model is purely video-based and requires no text input at the test phase, ensuring maximum flexibility and applicability. \KG{Ground truth for the test set is manually annotated.}}
    \label{fig:framework}
\end{figure*}

We present the \method\ framework for learning video OSCs, detailing the open-world formulation (Sec.~\ref{sec:problem_formulation}), model design for object-agnostic state prediction (Sec.~\ref{sec:model_design}), and text-guided training scheme (Sec.~\ref{sec:pseudo_label}). 

\subsection{Video OSC in the Open World} \label{sec:problem_formulation}
%\KGnote{early on give a clear definition of what counts as a state change (e.g., like in our annotation instructions)} \SXnote{added}
% While many approaches have touched on the topic of object state changes, there is no unified definition.   

% We begin by formally defining Object State Change (OSC) in videos: An OSC is a visually detectable transformation, where an object experiences a change that is not easily reversible. The definition aligns with ~\cite{grauman2022ego4d}, and we emphasize the two key attributes here: (1) visual detectability: OSCs should result in a visual change in the object's appearance. Processes that are non-visual or involve simple spatial transitions (e.g., moving an apple from the sink to the cutting board) are not considered OSCs; (2) state transition requirement: To qualify as an OSC, the video must present a visual transition from one state to another. Simply showcasing the end state (e.g., displaying sliced pineapples without depicting the slicing process) does not suffice. 
%\KGnote{this seems to contradict what we say above about object-centric; I think here you mean it must be seen both in original state and changed state, not just the changed state.} \SXnote{Yes, just to exclude the case where the video depicts only the end state, but not the state changing process. Seems like these two clarifications on OSC definition are not very related to our model, moved them to Supp.?}

We begin by formally defining an object state change (OSC) in videos as a visually detectable transformation, where an object experiences a change that is not easily reversible, in line with \cite{grauman2022ego4d}. We characterize an \emph{OSC category} as an object combined with a specific state transition, such as ``chicken + shredding'', and delineate an OSC process through three distinct states: initial \KG{(precondition state)}, transitioning, and end \KG{(postcondition state)}.\footnote{Due to real-world data variation, some videos may lack one of the OSC states, and there may be multiple segments in a video corresponding to the same state. Our formulation accounts for all these scenarios.} It is essential to highlight that we take an object-centric perspective: the ``transitioning state'' accounts for instances where the video depicts an active action applied to the object, as well as scenarios where the object undergoes a passive transformation absent of human intervention \KGCR{(e.g., melting, drying)}. 

\KG{Given a video in which an object may be changing state,} the objective is to \KG{temporally localize} %identify 
each of the three OSC states.  %within a given video, that is considered to contain an OSC.
 %\cc{The specific OSC category name associated with the video (e.g. shredding chicken) is available for use\KGnote{this is confusing} \SXnote{Fig. 2 (b) we use this OSC category name to generate the text query for pseudo labels}, but not directly provided as model input, since our goal is a vision-centric approach that learns OSCs without relying on text prompts for wider practical use.} %\KGnote{here it is unclear if we are assuming we know that there is a specific given state change to search for/localize in the input video.} \SXnote{revised} 
%\KGnote{at test time we output the name of the object and verb? or just localize the windows in time?} \SXnote{just localize the windows in time, same as all baselines; the multi-task version (in Supp.) can output the verb name in the meantime}
Consistent with prior work~\cite{souvcek2022look,souvcek2022multi}, we formulate the task as a frame-wise classification problem.
% \footnote{Given the sporadic presence of non-OSC-related background moments within a video, frame-wise classification offers a more precise formulation compared to temporal localization.} 
Formally, a video sequence is represented by a temporal sequence of $T$ feature vectors $\mathbf X=\{\mathbf x_1, \mathbf x_2, \cdots, \mathbf x_T \}$, where $T$ denotes the video duration. The goal is to predict the OSC state label for each timepoint, denoted by $\mathbf{Y}=\{\mathbf y_1, \mathbf y_2, \cdots, \mathbf y_T\}$, $\mathbf y_t \in \{1, \cdots, K+1\}$, where $K$ is the total number of OSC states, and there is one additional category representing the background class not associated with any OSC state.\footnote{\ccCR{Not all training videos may feature an OSC due to data collection noise, yet all evaluation videos are manually verified to include one.}} 
% \KGnote{In the Supp.~we report a multi-task setting where the model also outputs the transition type (e.g., ``shredding").} \SXnote{I think we need to delete this sentence, does the clarification in Sec. ~\ref{sec:experiment} implementation work? This formulation already accounts for multi-task variant, single task means K=3 all the time, and stating this here may look confusing}

Next, we propose an open-world problem formulation, capturing the intrinsic long-tail distribution observed in real-world scenarios. Consider $N$ state transitions, each paired with a set of associated objects, denoted by $\mathbf O^n = \{o_1^n, o_2^n, \cdots, o_{m(n)}^n\}$, for $n=\{1,2,\cdots, N\}$, where $m(n)$ denotes the number of objects linked with the $n$-th state transition. Within a specific state transition (like frying), certain objects in the set (such as chicken or onions) are frequently observed, while the combination of the same state transition with other objects (such as cauliflower) appear less often. Motivated by this inherent long-tail, we propose to split $\mathbf O^n$ into two \SX{distinct} subsets: $\mathbf O_{\text{known}}^n$ covering the common combinations observed during training, and $\mathbf O_{\text{novel}}^n$ comprising the infrequent combinations, which are unseen during training due to their rarity. 
% $\mathbf O_{\text{known}}^n \cup \mathbf O_{\text{novel}}^n = \mathbf O^n, \mathbf O_{\text{known}}^n \cap \mathbf O_{\text{novel}}^n = \varnothing$. 
During \KGCR{inference}, %test, 
the 
\KGCR{model is} %model's predictions are
evaluated on the entire object set $\mathbf O^n$ for a comprehensive reflection of the open-world setting.

\subsection{Object-Agnostic State Prediction} \label{sec:model_design}
% With the new problem formulation in hand, we now delve into methods tailored to address the open-world challenges and ensure robust generalization to novel objects. Central to our approach is the ability to learn generalized representations of states and state transitions that are not tied to specific objects. To this end, we introduce three techniques:
To address the open-world challenges and ensure robust generalization to novel objects, \method\ integrates three key techniques: (1) a shared state vocabulary that consolidates the state label space; %$K$; %this is a number
(2) temporal modeling within the video OSC model design; and (3) object-centric features. % to enrich feature representations.
% that enable the model to learn generalized object state representations not tied to specific objects. 

\custompar{Shared State Vocabulary} 
% Prior work~\cite{souvcek2022look,souvcek2022multi} typically assigns unique labels to states from different OSC categories, resulting a large label set with $K=3\times\sum_{n=1}^N m(n)$, where 3 denotes the number of OSC states (i.e., initial, transitioning and end) and $\sum_{n=1}^N m(n)$ is the total number of OSC categories. For instance, end states like melted butter and melted chocolate are treated as two distinct categories. This compartmentalized view can inadvertently hinder the model's generalization ability. Yet, on closer inspection, states across varied objects are inherently linked.  For instance, the ``melting'' principle remains consistent, \KG{even when applied to visually distinct objects} % even though visually, objects 
% like butter and chocolate.
% exhibit pronounced differences. 
\SX{The inherent link among different OSC categories is overlooked in prior work. Common practice involves developing separate models for each OSC category~\cite{alayrac2017joint,souvcek2022look} (e.g., one model exclusively for melting butter and another for melting chocolate), or using one single model that still treats every OSC as a distinct entity in the label space~\cite{souvcek2022multi} (e.g., considering melted butter and melted chocolate as two separate end states).}
% \KGnote{as written this sounds like the same thing.}\SXnote{revised}
Such a formulation results in an extensive label set with $K=3\times\sum_{n=1}^N m(n)$, where 3 denotes the number of OSC states (i.e., initial, transitioning and end) and $\sum_{n=1}^N m(n)$ is the total number of OSC categories. 
% \SXnote{I'm thinking if we should omit this sentence and the notation about $K=3\times\sum_{n=1}^N m(n)$ also for the next few sentences, this is meant for multi-task formulation, in single-task formulation, it would be a large number of individual models, not an expanded label space} 
This compartmentalized view can inadvertently hinder the model's generalization ability. Yet, on closer inspection, states across varied objects are intrinsically related.  For instance, the ``melting'' principle remains consistent, even when applied to visually distinct objects like butter and chocolate.
% This motivates us to group training of OSCs that share the same state transition together, and treat same-state labels \emph{across objects} as one to reduce the label set size to $K < 3 \times \sum_{n=1}^N m(n)$.
% This motivates us to group same-state labels \emph{across objects} as one label, thereby reducing the label set size to $K < 3 \times \sum_{n=1}^N m(n)$.
% \SXnote{in the single-task case, it would be group training of different OSC models that share the states together; in the multi-task case, it is reducing the label set size} 
\finalv{This motivates us to group OSC labels that share the same state transition as one, and train a single model per state transition.\footnote{See Supp.~for the multi-task model variant, where we develop one unified model for all state transitions.}}
% Consequently, the model is encouraged to learn common representations that characterize the initial, transitioning, and end states of ``melting'' that are shared by both butter and chocolate. This state sharing strategy not only simplifies \SX{training} % the label space 
% but also aids the model in transferring knowledge of state transitions from known to novel objects. 
\finalv{%Consequently, the adoption of 
\KGCR{By adopting} object-agnostic OSC labels, \KGCR{we} encourage the model to learn state representations shared by visually different objects, and thus facilitate the transfer from known to novel objects.}

\custompar{Temporal Modeling} Recognizing that state changes unfold over time, it is important to capture the temporal dynamics in videos. An object's state at any frame, be it initial, transitioning, or end, is often best understood in the context of preceding and succeeding frames. \SX{Contrary to prior works~\cite{alayrac2017joint,souvcek2022look,souvcek2022multi} that rely on isolated frame-wise modeling without considering the temporal progression of video OSCs, we address this gap by proposing a temporally-aware model design.}
% Hence, mere isolated frame-wise modeling, as employed by many prior works~\cite{alayrac2017joint,souvcek2022look,souvcek2022multi}, \KGnote{how absolutely can we state this?  does none of the prior work use temporal context? 
% this part of our design is straightforward, so if we can say more concretely it simply hasn't been done it will be stronger.}
% \SXnote{yes, none of them uses temporal context, just an image encoder to encode frame features, no temporal modeling.}  \KGnote{then say more strongly than ``as employed by many prior works"?} might fall short in capturing these intricate transitions. We thus propose a %more
% temporally-aware model design.
%\KGnote{it seems like here we are getting concrete about what the model is, but it's appearing only midway through 3.2. should we have an overall framework introduced first, and then explain the 3 key pieces that enable learning generalized representations?} \SXnote{revised the beginning sentence in Sec.\ref{sec:model_design}}
% An input video $\mathbf X=\{\mathbf x_1, \mathbf x_2, \cdots, \mathbf x_T \}$ is first encoded into a temporal sequence of feature representations $\mathbf Z=\{\mathbf z_1, \mathbf z_2, \cdots, \mathbf z_T \}$ and then decoded as a sequence of predicted OSC state labels $\tilde{\mathbf Y}=\{\tilde{\mathbf y}_1, \tilde{\mathbf y}_2, \cdots, \tilde{\mathbf y}_T \}$. For the encoding phase, we first employ a frozen video encoder~\cite{wang2022internvideo} $f_0$ to extract video features from $\mathbf X$, denoted by $\mathbf Z^0=\{f_0(\mathbf x_0), f_0(\mathbf x_1), \cdots, f_0(\mathbf x_T) \}$. 
As illustrated in Fig.~\ref{fig:framework}(c), we adopt an encoder-decoder architecture.
% We adopt an encoder-decoder architecture. \KG{See Fig.~\ref{fig:framework}(c).} 
For the encoding phase, input video features $\mathbf X=\{\mathbf x_1, \mathbf x_2, \cdots, \mathbf x_T \}$ are first projected using an MLP $f_{\text{project}}$, and augmented with sinusoidal positional embeddings, $\mathbf Z_{\text{pos}}$. Subsequently, a transformer encoder~\cite{vaswani2017attention} $f_{\text{transformer}}$ is adopted to capture the temporal dynamics among these features, yielding $\mathbf Z = f_{\text{transformer}}(f_{\text{project}}(\mathbf X)+\mathbf Z_{\text{pos}})$. For the decoding phase, a MLP decoder $g$, maps these temporally-aware hidden representations to OSC state predictions: $\tilde{\mathbf Y} = g(\mathbf Z)$. \KG{See Sec.~\ref{sec:experiment} for architecture and training details.} 
% \SXnote{added last sentence in Sec.~\ref{sec:pseudo_label} about training, model architecture is in exp section implementation. shall we omit this sentence here?} \KGnote{I'd keep it here}  
This design ensures that when predicting the state for a frame, the model has assimilated temporal context from the entire sequence.
% Consequently, the focus on object transformations along time---rather than object static appearance---enhances the model's ability to predict OSCs for novel objects.  \KG{In other words, our model exploits the fact that the \emph{dynamics} of the state change has greater invariance to the object category 
% than how the object looks.} \SXnote{what does the object category mean? I'm not sure I get this sentence} \KGnote{trying to say that we'll be better able to ignore the ways the object categories look different by paying attention to dynamics.}
% try to condense a bit
\SX{Essentially, our model exploits the fact that the \emph{dynamics} of the state change has greater invariance to the object category than how the object looks, emphasizing \KGCR{temporal} object transformations %along time 
over \KGCR{objects'} static appearances, \KGCR{and thereby enhancing} %to enhance
generalization %capabilities 
to novel OSCs.}

\custompar{Object-Centric Features} Finally, we discuss how to better represent ``object'' features \ccCR{in the problem}. In many scenarios, due to camera placement and framing, the object going through a state transition might only occupy a small portion of the frame, surrounded by other visual elements such as background, %individuals, 
\KG{people}, or bystander objects. Recognizing this challenge, we introduce an enhancement to our model's input features $\mathbf X$ to emphasize the object of interest. To be specific, we leverage an off-the-shelf detector~\cite{shan2020understanding} to identify the active object (i.e., the object being manipulated) region at each timepoint $t$, yielding feature $\mathbf x_t^{obj}$ that centers on the object (see bounding boxes in Fig.~\ref{fig:framework} (c)). The input feature is then constructed as a concatenation of the original global feature and the localized object-centric feature, i.e., $\mathbf X = \{[\mathbf x_t, \mathbf x_t^{obj}]\}_{t=1}^T$. By emphasizing the object in this manner, our model is better positioned to discern the intricate state changes and provide more informed predictions. 

\subsection{Text and VLMs as Supervision}  \label{sec:pseudo_label}
\SX{To scale up training and ensure broad generalization, we propose a novel training pipeline that leverages LLMs and VLMs for OSC mining and pseudo-labeling. 

\custompar{Mining for OSC examples} Utilizing a vast collection of \KGCR{``how-to"} instructional videos as the training source, we develop an automated mining process to capture the rich, real-world variability of OSCs. 
% This approach effectively identifies videos of interest (i.e., those containing OSCs) and concurrently develops an OSC vocabulary. 
The motivation is that instructional videos usually have accompanying \finalv{Automatic Speech Recognition (ASR)} transcriptions that offer valuable OSC cues. For example, a speaker %in the video
mentioning, ``so just get your rotisserie chicken and shred it up'' suggests that the chicken may undergo a state transition of shredding in the video. Leveraging this \KGCR{fact}, %finding, 
we employ LLAMA2~\cite{touvron2023llama2} to analyze ASR transcriptions for identifying candidate videos and their associated OSC categories. This text mining stage allows us to corral a long tail of OSCs, discovering likely state change terms---even if rare---from the free-form natural language narrations given by how-to instructors in the training videos. See Fig.~\ref{fig:framework} (a).}

\custompar{Pseudo Label Generation} Next, to negate the need of manually labeling large-scale training data, we propose a novel pseudo-labeling pipeline facilitated by VLMs. From the identified OSC category (e.g., shredding chicken), we form three \KG{\emph{textual state descriptions}} for its initial, transitioning, and end states (e.g., whole chicken, shredding chicken, and shredded chicken). We then adopt both the vision and language encoder from a well-trained VLM \SX{(we experiment with CLIP~\cite{clip} and VideoClip~\cite{xu2021videoclip})} %(we use CLIP~\cite{clip}) 
to compute the cross-modal similarity between every \KG{frame} %element 
in \KG{training video} % $\mathbf X$ 
and the three %text queries
\KG{state descriptions}, producing a score matrix $\mathbf S \in \mathbb{R}^{T\times 3}$. The pseudo label 
$\hat y_t$ at timepoint $t$ is then assigned based on this score matrix:

\finalv{
{\footnotesize\[
\hat y_t = 
\begin{cases} 
\text{Background} & \text{if } \SX{\sum}(\mathbf{S}[t, :]) < \tau \\
\text{Initial} & \text{elif } \mathbf{S}[t, 0] - \mathbf{S}[t, 1] > \delta \text{ and } \mathbf{S}[t, 0] - \mathbf{S}[t, 2] > \delta \\
\text{Transitioning} & \text{elif } \mathbf{S}[t, 1] - \mathbf{S}[t, 0] > \delta \text{ and } \mathbf{S}[t, 1] - \mathbf{S}[t, 2] > \delta \\
\text{End} & \text{elif } \mathbf{S}[t, 2] - \mathbf{S}[t, 0] > \delta \text{ and } \mathbf{S}[t, 2] - \mathbf{S}[t, 1] > \delta \\
\text{Ambiguous} & \text{otherwise}
\end{cases}
\]}
}where $\delta$ is the threshold that separates states and $\tau$ is the threshold that differentiates between states and background. \KG{Essentially if the VLM scores some state more strongly than the other two---\SX{and the cumulative confidence score of all states is high}% and stronger than the absolute confidence threshold $\tau$ 
---then it is adopted as the pseudo label. Otherwise it is omitted as ambiguous.} \KG{See Fig.~\ref{fig:framework}(b).} \SXnote{made a mistake in the equation, our implementation is sum rather than max}

To further refine these labels, we enforce a causal ordering constraint. Given the inherent progression of OSCs, the anticipated order is initial states followed by transitioning states, and finally, the end states. Any frame whose pseudo label does not respect this natural progression is re-assigned to the ambiguous category. \SX{Our training employs a cross-entropy loss between pseudo label $\hat y_t$ and the corresponding model prediction $\tilde y_t$, with ambiguous frames excluded to maintain clarity and distinction among the state labels.} \finalv{See Section \ref{sec:supp_result} in Supp. for detailed pseudo label analysis.}
 % All ambiguous frames are  masked out during training.
% \cc{These strategies help ensure a distinct separation between different states and the background, enhancing robustness of our pseudo labeling process.} 
\section{The \dataset\ Dataset} \label{sec:dataset}
%\KGnote{Do we want a separate section about the dataset construction?}
%\KGnote{better name/acronym?} \SXnote{a tentative name, feel free to replace it}

% Current datasets for studying OSCs are limited in their ability to represent the open-world complexity and diversity of OSCs. To address this gap, we introduce a large-scale dataset collection, \dataset, designed to facilitate open-world exploration of OSCs. In this section, we outline the methodology for our data and label collection.
Existing OSC datasets fall short in capturing the open-world's diversity and long-tail distribution of OSCs. To address this gap, we introduce \dataset. It encompasses varied state transitions coupled with a diverse range of objects, providing an authentic reflection of \KGCR{their} real-world frequency---from the commonplace to the rare. 
% \cc{Below, we detail our data and label collection process.}
%\KGnote{should we be clear up front that this means we want state changes that apply to many different objects, even if only rarely for some of them?} \SXnote{revised}

\begin{figure}[t!]
    \centering
    \includegraphics[width=1.0\linewidth]{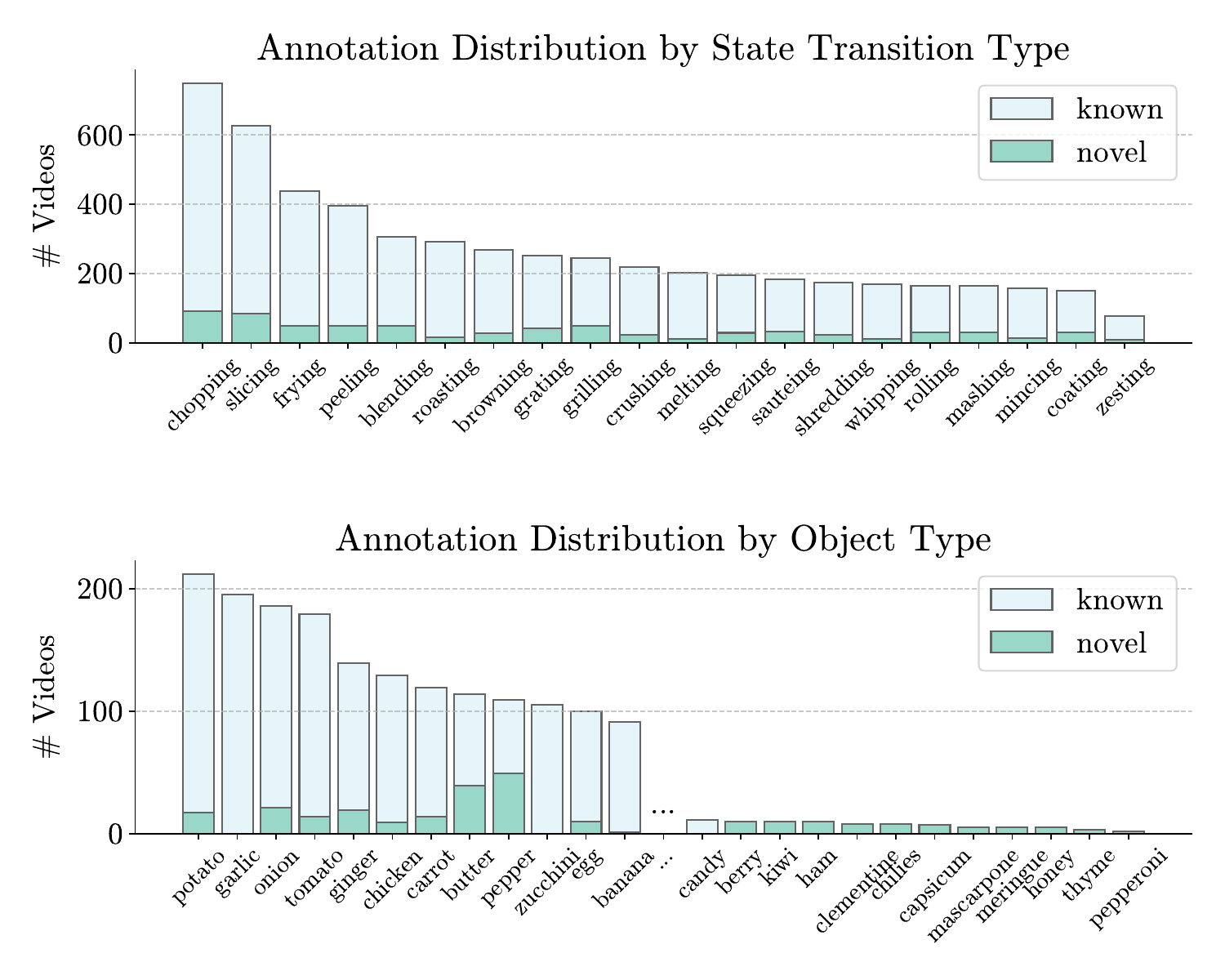}
    \vspace*{-5mm}
    \caption{\KG{Ground truth} annotation distribution across 20 state transitions (top) and 134 objects (bottom) in \dataset\ (Evaluation). In line with our open-world formulation, annotations cover a diverse range of object-state transition combinations, categorized into known and novel OSCs. }
    \label{fig:ann_distribution}
\end{figure}

\custompar{Data Collection} 
% To capture the rich, real-world variability of OSCs, we opt for in-the-wild YouTube videos rather than curated datasets collected in controlled environments. The HowTo100M collection~\cite{miech2019howto100m} is particularly suitable for this task, as it contains a wealth of instructional videos that often feature OSCs. 
% \KGnote{shouldn't this para go in 3.3 instead?} We design an automated process to not only acquire videos of interest (i.e., those containing OSCs) but also to simultaneously build an OSC vocabulary. The motivation is that instructional videos usually have accompanying ASR transcriptions that offer valuable OSC cues. For example, a speaker in the video mentioning, ``so just get your rotisserie chicken and shred it up'' suggests that the ``chicken'' is likely to undergo a state transition of ``shredding'' in the video. Leveraging this finding, we employ LLAMA2~\cite{touvron2023llama2} to process the ASR transcriptions and automatically identify segments within these videos that are likely to feature state changes based on the auto-aligned HowTo100M video-text pairs~\cite{han2022temporal}, as shown in Fig.~\ref{fig:framework} (a). 
The HowTo100M collection~\cite{miech2019howto100m} contains a wealth of instructional videos that often feature OSCs and is particularly suitable for this task. We specifically focus on the HowTo100M Food \& Entertaining category \KG{because (1) it constitutes a third of the entire HowTo100M videos, (2) cooking tasks offer a wealth of objects, tools, and state changes, providing an excellent test bed for open-world OSC, and (3) in cooking activities a single state transition can often be associated with a varied range of objects, opening the door to learning compositionality.  (Note, we also experiment with non-cooking domains below.)}
We process a total of 498,475 videos and 11,390,287 ASR transcriptions %\KGnote{what are the text queries?} \SXnote{revised}
%from HowTo100M Food \& Entertaining category 
with LLAMA2.  
%as (1) it constitutes roughly a third of the entire HowTo100M videos; (2) cooking presents a domain where objects commonly undergo changes, and a single state transition can be associated with a varied range of objects; (3) culinary tasks are generally of lower complexity, making them more suitable for annotation by non-experts.}} with LLAMA2. 
From the responses, we identify the most frequently seen state transitions and objects associated with them to establish an OSC vocabulary, resulting in 134 objects, 20 state transitions, and 409 unique OSCs. The number of objects associated with each state transition \KG{($m(n)$)} spans from 6 for ``zesting'' to 55 for ``chopping''. The state transitions applied to each object vary from 1 to 15, with onions being the most versatile. \SX{In total, we identify 36,075 videos for the training set of \dataset, with an average duration of 41.2 seconds.}
% displaying the greatest versatility, being involved in 15 different state transitions. 

%\KGnote{explain this arithmetic.  like every one of the 134 objects has some number of possible transitions that apply to it (even if sometimes only very rarely) and...} \SXnote{added} 
% This innovative approach to training data collection greatly reduces manual effort, while effectively capturing real-world OSC distributions.  

% It is worth noting that the acquired videos come with their share of noise. Due to limitations in LLAMA2's performance and potential misalignment between video and text, not all identified segments may accurately contain the corresponding OSC. However, we consider this inherent noise to be an asset for our goals, as it adds a layer of real-world complexity, aiding in the training of a more robust video OSC model at scale.  
%\KGnote{be clear that once we evaluate, it is with manual annotations.} \SXnote{added in footnote in label collection paragraph. moved sentences starting from ``It is worth noting..'' to Supp.}
% See Supplementary for the LLAMA2 text prompts and more data collection details. 

\custompar{Data Splits}
In line with our open-world formulation, we divide the 409 identified OSCs into two disjoint subsets based on their frequency of occurrence.  Within each state transition, we categorize the top 75\% frequent objects as known and the bottom 25\% as novel. 
% \cc{Consequently, commonly observed OSC combinations, like ``whipping cream'' fall under known categories and are available during training while rarer combinations such as ``whipping strawberry'' are withheld from the training set to test models' generalization capability.}
% are withheld from the training set and only introduced during testing to evaluate the model's capacity for generalization. 
% Utilizing a 0.25 quantile threshold for each state transition, we categorize the top 75\% frequent OSCs as known and the bottom 25\% as novel. 
This yields a total of 318 known OSCs \KG{that are seen during training and testing}, spanning 20 state transitions associated with 122 objects, and 91 novel OSCs \KG{that are only seen during testing}, encompassing the same transitions across 58 objects.
% A detailed OSC taxonomy is available in Supplementary. 

\custompar{\KG{Ground Truth} Label Collection \KG{(Evaluation Set)}} 
To facilitate thorough evaluation, we obtain manual annotations for %annotate 
a subset of 5,424 videos from the collected dataset.\footnote{Our training is purely guided by a VLM and requires no ground truth labels. The annotations are reserved exclusively for evaluation.} The annotation workflow is as follows: each annotator is presented with a video segment along with an OSC category that was previously identified by LLAMA2. The annotator has the option to reject the video segment if it does not contain the specified OSC. Otherwise, they label the time ranges corresponding to the initial, transitioning, and end states of the OSC. \SX{Adhering to our object-centric emphasis, annotators are instructed to label based on the visual changes of the object, rather than human-centric actions, and exclude time ranges where the object of interest is not visible, ensuring clean and focused temporal labels.} %\SXnote{condensed the original paragraph}
Fig.~\ref{fig:ann_distribution} provides the distribution of annotated videos.
% , detailing the frequency across 20 state transition types and 134 object types in \dataset. %\KGnote{trying to understand why 122 objects in train in table.} \SXnote{object number associated with known OSCs, added this number in last sentence of data splits paragraph}  
On average, we collect 271 annotations per state transition, \SX{with a video duration of 41.5 seconds}, and 12.9\% of videos belong to the novel category.
The entire annotation process required around 1,507 hours by 30 professional annotators.
% It's also noteworthy that an object (e.g., potato and onion) can be associated with multiple state transitions, encompassing both known and novel categories. Intriguingly, there are objects (e.g. kiwi and clementine) that never feature in the training set, posing the challenge of recognizing their linked OSCs during evaluation. 

\begin{table}[!t]
\tablestyle{1.5pt}{1.2}
  \centering
  \begin{tabular}{lcccccc}
    \thickhline
    Datasets & \# Obj & \# ST & \# OSC & \makecell[c]{\KGCR{ObjPer}} & \# Videos & \makecell[c]{\SX{GT}\\ Label?} \\
    \hline
    % MIT-States~\cite{mit_states} & 245 & 115 & 2550 & 5.3K images &  \\
    % Ego4D~\cite{grauman2022ego4d} & \\
    Alayrac et al.~\cite{alayrac2017joint} & 5 & 6 & 7 & 1.2 & 630 & \greencheck \\
    Task-Fluent~\cite{liu2017jointly} & 25 & 14 & 32 & 2.3 & 809 & \greencheck \\ 
    ChangeIt (Training)~\cite{souvcek2022look} & 42 & 27 & 44 & 1.6 & 34,428 &  \redcross \\
    ChangeIt (Evaluation)~\cite{souvcek2022look} & 42 & 27 & 44 & 1.6 & 667 & \greencheck \\
    % ChopNLearn~\cite{saini2023chop} & 20 & 8 & 112 & 14.0 & 1,260 & \ding{55}\\
    % VSCOS~\cite{yu2023video} & 124 & 30 & 271 & 9.0 & 1,905 & \ding{55} \\
    \hline
    % \rowcolor{LighterGray}
    \dataset\ (Training) & 122 & 20 & 318 & 15.9 & 36,075 & \redcross \\ 
    % \rowcolor{LighterGray}
    \dataset\ (Evaluation) & 134 & 20 & 409 & 20.5 & 5,424 & \greencheck \\
    \thickhline
  \end{tabular}
  \caption{Comparison with existing video datasets focusing on object states. `Obj' and `ST' represent objects and state transitions, respectively. `\KGCR{ObjPer}' denotes the average number of objects associated with \KGCR{each} %a 
  state transition; \KG{higher values indicate more need to generalization across objects}. We present the first open-world benchmark for temporal video OSC, with an order of magnitude increase in OSC vocabulary and annotation volume.}
  \label{tab:dataset}
\end{table}

\custompar{Dataset Comparison}
Table \ref{tab:dataset} compares \dataset\ with existing video datasets on \SX{temporal OSC understanding}. \dataset\ offers an unprecedented scale---with 9.3x more OSC categories and 8.1x more annotated video instances compared to the previous largest collection~\cite{souvcek2022look}. \KGCR{Furthermore}, notably, in prior datasets, 
% are constrained in data and label scale, where 
each state transition is typically coupled with 1 or 2 objects, \KGCR{preventing subsequent models from generalizing to new objects,} %. Consequently, models developed from these datasets are likely limited to recognizing OSCs for the few objects they have been exposed to and unable to generalize to new objects, 
\KG{as we will see in results.}
In contrast, \dataset\ pioneers %an innovative 
the open-world formulation, presenting a much broader range of objects associated with each state transition\KGCR{---from 6 to 55 objects per state transition, and averaging 20.  This facilitates} the development of models with %more 
generalized OSC understanding.
% overlook the compositional nature between objects and state transitions, In contrast, \dataset\ pioneers an innovative open-world formulation, presenting the most expansive dataset and richest annotation collection in this field. 
Please see Supp. for full data collection and annotation details.
% or lack temporal annotations required for our task.
% Supp: duration & segment statistic
\section{Experiments} \label{sec:experiment}

\begin{table*}[!t]
\tablestyle{4pt}{1.1}
  \centering
\begin{tabular}{l|cc|cccc|cccccc}
 \thickhline
     \multirow{3}{*}{Method} & \multicolumn{2}{c|}{ChangeIt} &  \multicolumn{4}{c|}{ChangeIt (open-world)} & \multicolumn{6}{c}{\dataset} \\
    & \multirow{2}{*}{\makecell[c]{State \\Prec.@1}} & \multirow{2}{*}{\makecell[c]{Action \\Prec.@1}}& \multicolumn{2}{c}{State Prec.@1} & \multicolumn{2}{c|}{Action Prec.@1} & \multicolumn{2}{c}{F1 (\%)} & \multicolumn{2}{c}{Prec (\%)} & \multicolumn{2}{c}{Prec.@1 (\%)} \\
    &  &  & known & novel & known & novel & known & novel & known & novel & known & novel \\
\hline
CLIP~\cite{clip} & 0.30 & 0.63 & 0.29 & 0.29 & 0.71 & 0.70 & 26.9 & 25.4 & 27.3 & 26.6 & 47.5 & 47.5 \\
VideoCLIP~\cite{xu2021videoclip} & 0.33 & 0.59 & 0.25 & 0.24 & 0.62 & 0.55 & 36.6 & 34.3 & 39.7 & 38.5 & 48.3 & 44.8 \\
InternVideo~\cite{wang2022internvideo} & 0.27 & 0.57 & 0.29 & 0.25 & 0.60 & 0.61 & 29.9 & 29.5 & 31.4 & 30.8 & 46.9 & 46.3 \\
LookForTheChange~\cite{souvcek2022look} & 0.35 & 0.68 & 0.36 & 0.25 & 0.77 & 0.68 & 30.3 & 28.7 & 32.5 & 30.0 & 37.2 & 36.1 \\
MultiTaskChange~\cite{souvcek2022multi} & 0.49 & 0.80 & 0.41 & 0.22 & 0.72 & 0.62 & 33.9 & 29.9 & 38.5 & 34.1 & 43.1 & 38.8 \\
\rowcolor{LighterGray} 
% \method\ (multi-task) & 0.44 & 0.69 & 0.43 & 0.29 & 0.75 & 0.63 &  40.7 & 37.9 & 41.8 & 39.0 & 56.8 & 54.8 \\
\method\ (ours) & \textbf{0.57} & \textbf{0.84} & \textbf{0.56} & \textbf{0.48} & \textbf{0.89} & \textbf{0.82} & \textbf{46.4} & \textbf{43.1} & \textbf{46.6} & \textbf{43.7} & \textbf{60.7} & \textbf{58.2} \\
 \thickhline
\end{tabular}
\caption{Results on ChangeIt, ChangeIt (open-world), and \dataset. \method\ outperforms all approaches in both closed-world and open-world scenarios, across known and novel OSCs.}
\label{tab:combined_results}
\end{table*}

\begin{figure*}[!t]
    \centering
    \includegraphics[width=0.95\linewidth]{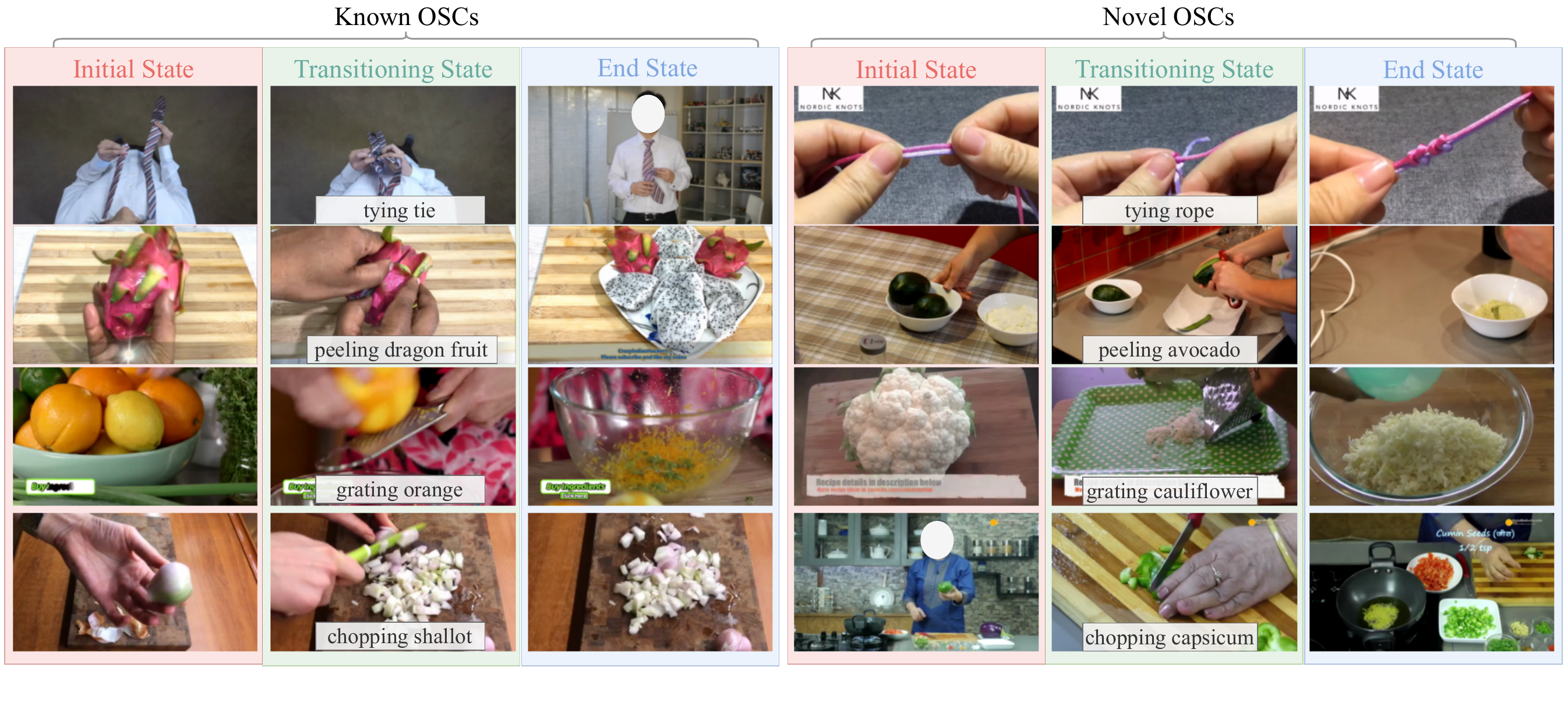}
    \vspace*{-3mm}
    \caption{Top-1 frame predictions given by \method\ for the initial, transitioning, and end states, on ChangeIt (open-world) (first 2 rows) and \dataset\ (last 2 rows). \method\ not only accurately \KG{localizes} %identifies
    the three fine-grained states for known OSCs, but also generalizes this understanding to novel objects, such as cauliflower and capsicum, which are not observed during training. \SXnote{updated the figure}}
    \label{fig:viz_prediction}
\end{figure*}

% \subsection{Experimental Setup} \label{sec:setup}

\custompar{Datasets}
In addition to our new \dataset\ dataset, we also evaluate on ChangeIt~\cite{souvcek2022look} due to its expansive data scale (% \KG{2600 hours of video spanning many activities} 
\SX{34K videos spanning many activities}) \SXnote{try to avoid mentioning their video duration because our video duration is shorter than theirs}
and high relevance to our task.   %\footnote{As outlined in Table \ref{tab:dataset}, the small OSC vocabulary and dataset size renders other relevant datasets~\cite{alayrac2017joint,liu2017jointly} unsuitable for our context. \KGnote{I don't think this is convincing.} \SXnote{do not have a better idea of how to state this. maybe omit this sentence and describe in Supp. that the OSC vocabulary in ~\cite{alayrac2017joint} involves ``open'' and ``close'', not very aligning with our adopted OSC definition (not easily reversible)? dataset~\cite{liu2017jointly} are not publicly released.}}
Beyond the conventional split of ChangeIt, we propose a new split tailored to our open-world formulation. Specifically, from the 44 available OSC categories in ChangeIt, we concentrate on the 25 categories where each state transition is paired with more than one object. %This has led to the creation of our distinct 
With those, we form the ``ChangeIt (open-world)" subset that comprises 8 state transitions and 25 objects. Within each state transition, objects are randomly divided into known and novel categories, yielding 13 known OSC categories for training and 12 novel OSC categories exclusively reserved for evaluation. %\cc{As an illustration, consider the ``melting'' transition which, in ChangeIt, relates to both butter and chocolate. In our division, ``melting butter'' is designated as the known OSC, whereas ``melting chocolate'' is categorized as novel (meaning videos of melting chocolate are omitted from training and only presented during evaluation). 
A detailed breakdown of this split can be found in Supp. To sum up, our evaluation encompasses: ChangeIt; ChangeIt (open-world); and \dataset, offering a comprehensive setup in both closed-world and open-world scenarios. %\KGnote{be sure to have a clear statement about why the other datasets we cite about state changes are not applicable here or why we choose only these} \SXnote{added}

\custompar{Evaluation} 
For ChangeIt and ChangeIt (open-world), we adhere to the original dataset's evaluation protocol~\cite{souvcek2022look}, reporting action and state precision@1 as the evaluation metrics. For our dataset, besides precision@1, which evaluates a single frame for each state within a video, we advocate for the use of F1 score and precision over all frames to ensure a more holistic evaluation. 
% \cc{We present these metrics on both known and novel OSCs for the two open-world datasets.} 

% For ChangeIt and ChangeIt (open-world), we adhere to the original dataset's evaluation protocol, reporting action and state precision@1 as the evaluation metrics. For our dataset, while we also adopt state precision@1, due to our definition that positions the midpoint between the initial and end states as ``transitioning states'' rather than action (emphasizing an object-centric perspective), our evaluation takes into account three distinct states, unlike the two states in ChangeIt. In addition, since precision@1 solely evaluates a single frame for each state within a video, we advocate for the use of F1 score and precision over all frames to ensure a more holistic evaluation. For each video, we compute the state precision@1, F1 score and precision for the present states (considering that a video might not always contain all three states: initial, transitioning, and end) and then compute their average over states. Subsequently, we average these values across videos within a state transition category and report the overall average for all state transitions. Lastly, for the two open-world datasets, we present these metrics on both known and novel OSC categories. 
% This process is treated as a 4-class classification, and the results are averaged across the three states to exclude the background. 

\custompar{Baselines} We compare our approach with four baselines across two categories: (a) self-supervised approaches on identifying object states enforced by the causal ordering constraint: LookForTheChange~\cite{souvcek2022look} trains a dedicated model for each OSC category, while MultiTaskChange~\cite{souvcek2022multi} evolves this into a multi-task approach\footnote{To ensure a thorough evaluation, we train both single-task and multi-task variants of our approach. See Supp. for a detailed discussion. %\SXnote{Moved all multi-task results and discussion to Supp.}
}, catering to several OSCs concurrently; (b) zero-shot VLMs: image-based CLIP~\cite{clip}, video-based VideoCLIP~\cite{xu2021videoclip} and InternVideo~\cite{wang2022internvideo}.
\KG{All baselines in (a) use the same training data as our model, whereas the zero-shot models (b) are directly evaluated on the test set with no training.}
%\KGnote{for seen object case, are there not more models from literature we should include for comparisons?} \SXnote{category (a): no more recent baseline on this topic besides these 2; category (b): added InternVideo baseline} 

\custompar{Implementation} Videos are sampled at one frame per second. Each one-second video segment gets assigned to an OSC state label, and is encoded by InternVideo~\cite{wang2022internvideo}, a general video foundation model (which is kept frozen for training efficiency). Our video model consists of a 3-layer transformer with a hidden dimension of 512 and 4 attention heads as the encoder and a 1-layer MLP as the decoder.
% \cc{We employ the AdamW optimizer with a learning rate of 1e-4 and a weight decay of 1e-4.}
% \cc{Models are trained using a batch size of 64, over 50 epochs.} 
Consistent with prior work~\cite{souvcek2022look,souvcek2022multi}, our model predicts the OSC state label (i.e., initial, transitioning, end state, or background) for a video, \finalv{assuming the video OSC category is known (say from a recognition model's output or a user's specific query).}
% within a given video. 
While the standard output does not include \KGCR{the} state transition name, our multi-task version detailed in Supp.~is capable of \KGCR{this}. %See Supp.~for details.
% further discussion on this and complete evaluation, baseline and implementation details.
% See Supp. for complete 

% \subsection{Results}

\custompar{Main Results} Table \ref{tab:combined_results} presents results on \KGCR{all} %the
three datasets. \method\ outperforms all approaches in both traditional closed-world and our newly introduced open-world settings. The large performance gains\KGCR{---as much as 9.6\% jumps in precision vs.~the next best method---}underscore the effectiveness of two pivotal components in \method. First, its use of text and VLMs for supervisory signals during training: % greatly boosts performance. P
\KGCR{particularly} on novel OSCs 
% \ccCR{not encountered during training}
, \method\ extracts meaningful cues from the video modality to refine pseudo labels from VLMs, and ultimately surpasses the VLM baselines. 
Second, our model for object-agnostic state prediction, designed for the open-world context, effectively narrows the gap between known and novel OSCs. For instance, on ChangeIt (open-world), MultiTaskChange~\cite{souvcek2022multi} experiences a 19\% decline in state precision@1 while \method\ has only a 8\% drop. Note that \SX{we observe a more pronounced performance drop of all approaches on ChangeIt (open-world)}
% the \KG{baseline's??} performance drop on ChangeIt (open-world) is more pronounced 
than on \dataset\ due to its limited number of objects available per state transition. These results also point to \KGCR{the} potential for future development in bridging the known and novel OSC performance gap. %SXnote{condensed the results paragraph}

\begin{figure}[!t]
    \centering
    \includegraphics[width=1.0\linewidth]{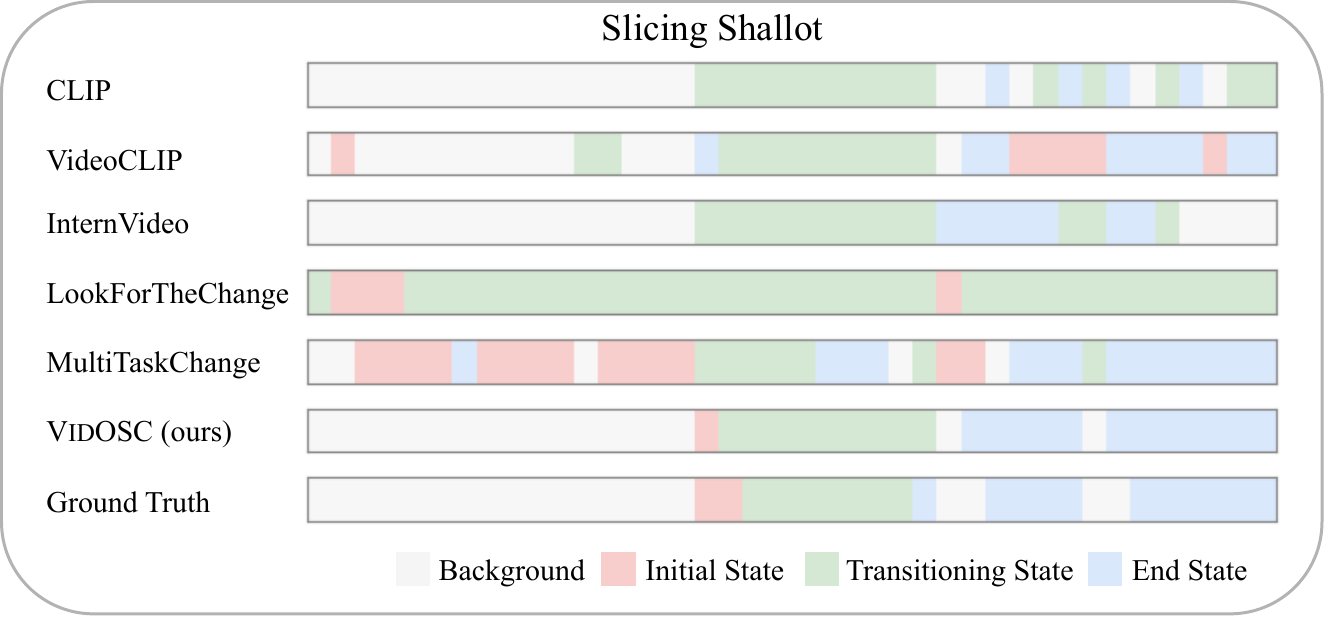}
    \caption{Comparison of model predictions across a test video depicting the OSC of ``slicing shallot'' on \dataset. The x-axis represents temporal progression through the video. \method\ gives temporally smooth and coherent predictions that best align with the ground truth, significantly outperforming baselines in capturing the video's global temporal context. %\KGnote{how to convince this isn't hand picked? is it typical?} \SXnote{yes it is typical, I can show more examples in Supp. Does it work?} YEs
    }
    \label{fig:viz_temporal_prediction}
\end{figure}

\custompar{Qualitative Results} Fig.~\ref{fig:viz_prediction} presents \method's top-1 frame predictions on ChangeIt (open-world) and \dataset, across both known and novel OSCs. Notably, despite never seeing objects such as cauliflower and capsicum during training, \method\ effectively leverages visual state change cues and correctly localizes the three states of these objects going through OSCs, demonstrating its strong generalization capability. 
% For each video, we showcase the frame with the highest probability assigned by \method\ for each of the three OSC states: initial, transitioning, and end.
% Notably, although \method\ never encounters the three objects during training, it accurately discerns the three fine-grained OSC states, based solely on the visual content. 
%Transitioning to
\KGCR{For} a more holistic view, Fig.~\ref{fig:viz_temporal_prediction} compares \method's frame-by-frame predictions with all baselines for a given test video. \method\ provides temporally coherent predictions, smoothly progressing through the OSC states in the natural order (i.e., from initial to transitioning then end). In contrast, baseline approaches often yield fragmented and inconsistent predictions, indicating a lack of understanding of the video’s global temporal context, primarily due to their reliance on frame-wise modeling. See Supp. and Supp. video for more qualitative examples and \method's interpretability on object relations. 

\begin{table}[!t]
\tablestyle{6pt}{1.2}
  \centering
  \begin{tabular}{cccccc}
    \thickhline
  Shared & Temporal & Object & \multicolumn{3}{c}{Prec.@1 (\%)} \\
 State & Modeling & Centric & known & novel & $\Delta$ \\
    \midrule
    \redcross & \greencheck & \greencheck & 58.5 & 53.3 & 5.2 \\
    \greencheck & \redcross & \greencheck & 52.9 & 48.2 & 4.7 \\
    \greencheck & \greencheck & \redcross & 59.8 & 56.7 & 3.1 \\
    % \hline
    \rowcolor{LighterGray}
  \greencheck & \greencheck & \greencheck & \textbf{60.7} & \textbf{58.2} & \textbf{2.5} \\
    \thickhline
  \end{tabular}
  \caption{Ablation Study. $\Delta$ denotes the performance gap between known OSCs and novel OSCs.}
  \label{tab:ablation}
\end{table}

\custompar{Ablation} To further dissect the performance gains brought by our three model design techniques (i.e., shared state vocabulary, temporal modeling, and object-centric features), we conduct an ablation study, removing one component at a time. %The results, presented in 
\KGCR{Table \ref{tab:ablation} confirms} the essential role of each element. A shared state vocabulary is particularly crucial in the open-world context, as its absence increases the gap between known and novel OSCs from 2.5\% to 5.2\%. Furthermore, temporal modeling provides a substantial performance boost, and object-centric features offer further gains. See Supp.~for an additional analysis of \method’s performance with different pseudo labels.

\begin{figure}[!t]
    \centering
    \includegraphics[width=1.0\linewidth]{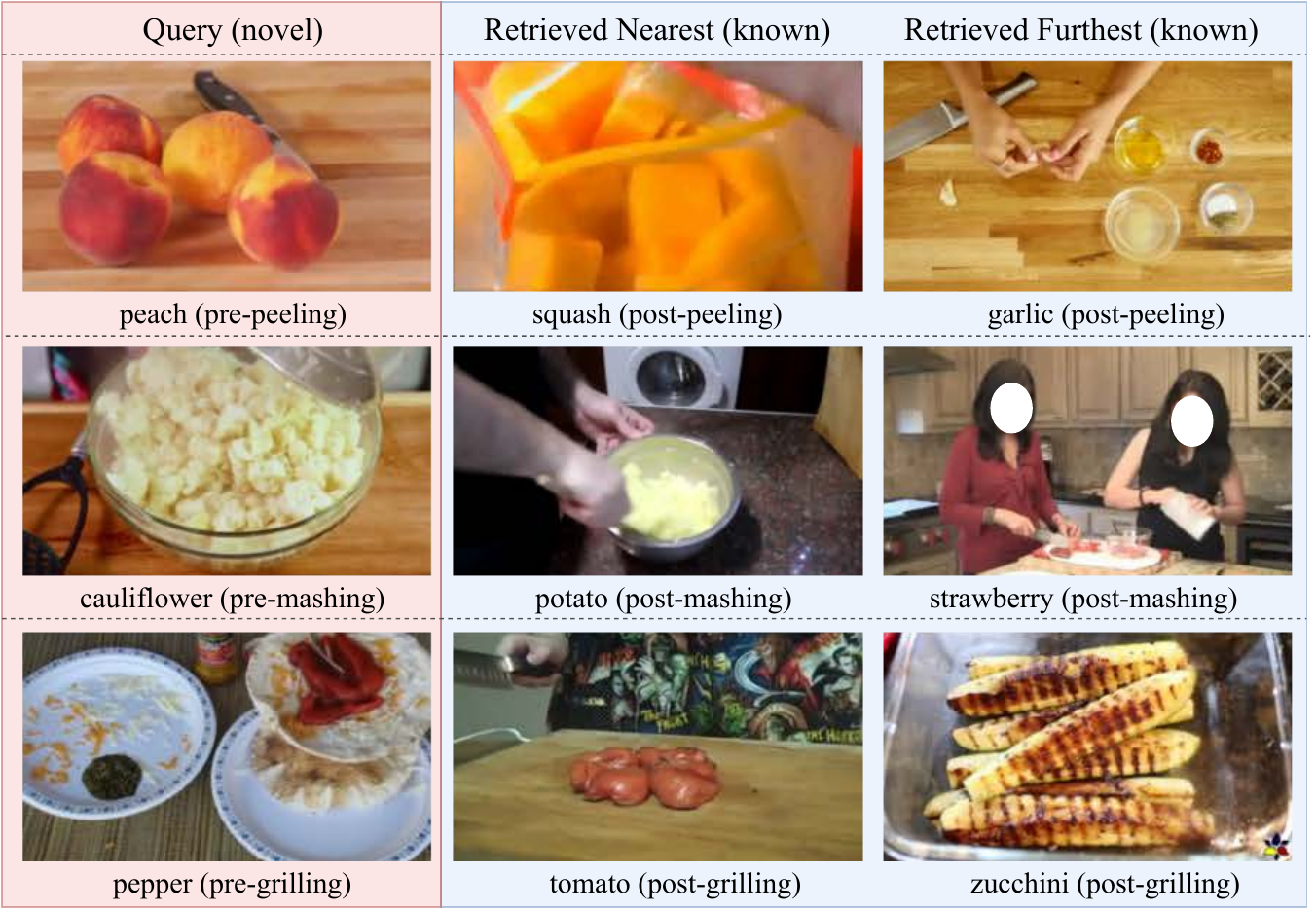}
    \caption{Frame retrieval on the \dataset\ test set. Given a query frame showcasing a \emph{novel} OSC at its \emph{initial} state, \method\ retrieves the nearest and furthest frame among all \emph{known} OSCs at their \emph{end} states. The closest post-OSC frame follows the transition trajectory of the query while the furthest post-OSC frame depicts a substantially different object, demonstrating \method's capability to generalize the OSC progression for novel objects. %\KGnote{would it be even more convincing to also retrieve the frames for the same state change, but different object, that is furthest from the query?} \SXnote{updated the figure}
    }
    \label{fig:viz_retrieval}
\end{figure}

\custompar{Frame Retrieval} \method\ learns features that accurately characterize 
% the three OSC states and effectively bridge the known and novel OSC gap. 
the evolution of an OSC process.
To illustrate, we consider a novel frame retrieval setting. Within the \dataset\ test set, when presented with a query video featuring a novel OSC at its initial state, \method\ seeks the most similar and most contrasting video from a pool of candidates at their end states. The frame triplets with the smallest and largest feature distance are shown in Fig.~\ref{fig:viz_retrieval}. Remarkably, the retrieved nearest post-OSC frames correspond to the query's anticipated state transition trajectory, despite the object and state gap. Conversely, the furthest frames exhibit end states of markedly different objects. These results further lend support to \method's capability to understand the evolution of an OSC process, even for novel objects it has never encountered during training.

%\KGnote{evidence of the object category relatedness helping to predict impact of new compositions? }
%\KGnote{long-tail figure?}
%\KGnote{illustrate how precondition appearance helps us guess the appearnace of the %object post-condition.  e.g., the whole egg appearance helps expect what it will look like once chopped.} 

\section{Conclusion}

This work aims at a comprehensive exploration of video OSCs, with a novel open-world formulation. 
% \cc{and an object-centric perspective}. 
To address the challenges, we leverage text and VLMs to assist the training of a video OSC model at scale and design three modeling techniques to achieve object-agnostic state prediction for better generalization to novel OSCs. Furthermore, we present the most expansive video OSC dataset collection \dataset, \KGCR{which} echoes the natural long-tail of state transitions coupled with varied objects, fostering a realistic representation of real-world scenarios. As for future work, we \KGCR{will} consider extending \method\ to %longer and more complex 
video sequences featuring concurrent OSC processes, and integrating spatial understanding of OSC within our open-world framework.

\small{\KGCR{\noindent \textbf{Acknowledgements:} UT Austin is supported in part by the IFML NSF AI Institute. KG is paid as a research scientist by Meta.}}

% Finally, we hope that our pioneering open-world formulation and the comprehensive \dataset\ coming with dense frame-wise annotations, shed light on deep OSC understanding in videos and facilitate research progress in this field and related areas.
\clearpage
{
    \small
    \bibliographystyle{ieeenat_fullname}
    \bibliography{main}
}
\clearpage
% \cleanfalse
\setcounter{page}{1}
\setcounter{section}{0}

\newif\ifarxiv
\arxivtrue

\ifarxiv
\section{Video Containing Qualitative Results}
We invite the reader to view the video available at \url{https://vision.cs.utexas.edu/projects/VidOSC/}, where we provide: (1) a comprehensive overview of \method, (2) video examples from \dataset, and (3) qualitative examples of \method's predictions. These examples highlight \method's ability in identifying non-OSC moments as background and effectively distinguishing among the three OSC states. It delivers temporally smooth and coherent predictions that follow the natural OSC progression (from initial to transitioning, and then to the end state), and shows strong performance even with novel OSCs not seen in training. All these underscore the efficacy of \method.

\else
\maketitlesupplementary
\section{Supplementary Material}
The supplementary materials consist of:
\begin{itemize}
\item A supplementary video offering a comprehensive overview of \method\ along with qualitatiie examples.
\item \SX{Additional explanations on video OSC definitions.}
\item Detailed information on \dataset, elaborating on the data and label collection process.
\item \SX{Experimental details: this includes complete experimental setup, further results and visualizations that are referenced in the main paper.}
% Experiments: This includes complete experimental setup, further experimental results \KG{mentioned in the main paper} and visualizations.
% \item An annotation guideline that we provide to the annotators for labeling the \dataset\ (Evaluation) set.
\end{itemize}

\section{Supplementary Video}
In the supplementary video, we provide: (1) a comprehensive overview of \method, (2) video examples from \dataset, and (3) qualitative examples of \method's predictions. These examples highlight \method's ability in identifying non-OSC moments as background and effectively distinguishing among the three OSC states. It delivers temporally smooth and coherent predictions that follow the natural OSC progression (from initial to transitioning, and then to the end state), and shows strong performance even with novel OSCs not seen in training. All these underscore the efficacy of \method.
\fi

\KGnote{somewhere - since we are citing Ego4D as the definition of state change, but then don't compare to it - can we make a clear statement about why that dataset is not applicable to our problem?  e.g. we're citing Ego4D in the first couple video slides.} \SXnote{added in experimental setup first paragraph}

\KGnote{I like to point out the lines in the main paper that each part below corresponds to, i.e., make it clear esp on results, where we had already foreshadowed that this would be provided.} \SXnote{added}

\section{Video OSC}
Expanding on Sec.~\ref{sec:problem_formulation}, we clarify three aspects of our definition of video OSC. First, we focus on OSCs that lead to a visible change in an object's appearance. Processes that are non-visual or involve mere spatial movements (such as moving an apple from the sink to the cutting board) do not qualify as OSCs. Second, in line with previous works~\cite{alayrac2017joint,liu2017jointly,souvcek2022look,souvcek2022multi}, we operate under the assumption that each input video predominantly features a single OSC. The challenge of handling videos with multiple concurrent OSCs remains an intriguing avenue for future research. Lastly, during training, the input video and its OSC category name (e.g., shredding chicken) are available (as provided in both ChangeIt and our \dataset, although not always accurate due to data collection noise). For evaluation, every test video (in both ChangeIt and \dataset) is accompanied by a manually verified OSC category.

\subsection{Data Collection}
We streamline our dataset collection via an automated process. First we apply LLAMA2~\cite{touvron2023llama2} to ASR transcriptions in HowTo100M~\cite{miech2019howto100m} with the following text prompt, one sentence at a time:

\begin{quote}
    \emph{[Text Prompt to LLAMA2]} You will receive descriptions corresponding to a how-to instructional video. Your task is to identify any instances of Object State Change (OSC) based on the provided text. An OSC is a visually detectable transformation where an object undergoes a change that is difficult to reverse. Examples include apple peeling/cutting, bacon frying, milk boiling, butter melting, cake frosting, eggs whisking, cream whipping, etc.

    \begin{itemize}
        \item Note 1: OSCs must be visually detectable. General actions like food preparing, or non-visual processes like onions sweetening, are not included.
        \item Note 2: Simple spatial transitions, resulting from actions like add, mix, put, or place, are not considered OSCs.
        \item Note 3: To qualify as an OSC, there must be a transition from one state to another, which should be indicated by an active action in the text description. The mere presence of a state (e.g., sliced pineapples, peeled apples) does not count unless there is explicit text describing the change (e.g., we slice the pineapples).
        \item Note 4: Generally, sentences without OSCs are far more common than those with OSCs.
    \end{itemize}

    To report identified OSCs, please use the following format: [object] + [state transition of the OSC]. Ensure the first word in each identified OSC is the object, and the subsequent words describe a state transition. If multiple OSCs are identified, separate them with semicolons (;). If no OSCs are detected, simply reply with None.
\end{quote}

\begin{figure}[!t]
    \centering
    \includegraphics[width=0.8\linewidth]{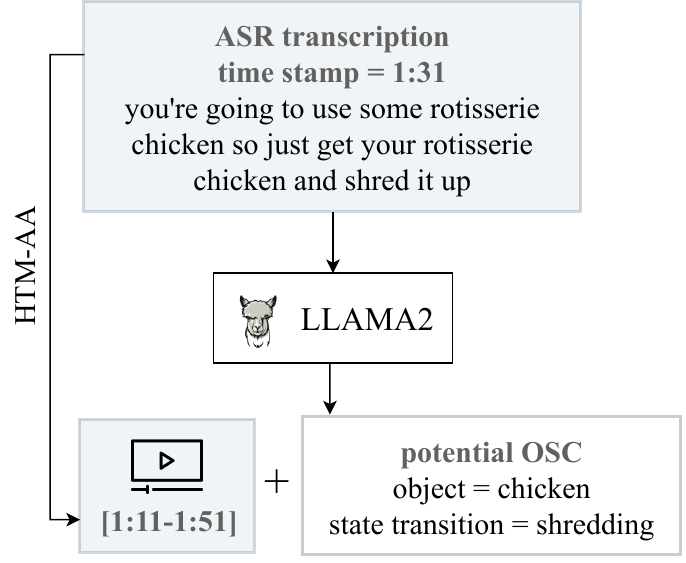}
    \caption{Our proposed data collection process for \dataset. HTM-AA~\cite{han2022temporal} denotes the auto-aligned version of HowTo100M.}
    \label{fig:data_collection}
\end{figure}

From the responses given by LLAMA2, we identify object and state transitions corresponding to the ASR transcription. Utilizing the HTM-AA dataset~\cite{han2022temporal}, where each ASR sentence corresponds to a time stamp in the video, we extract a clip centered around the identified time stamp with a $\pm$ 20 second window. The result is a cropped video segment of 40 seconds, paired with an OSC text (in the form of object + state transition). Finally, when multiple clips from the same video illustrate the same OSC with overlapping start and end times, we combine them into a single, extended clip. The whole process is illustrated in Fig.~\ref{fig:data_collection}.

\section{The \dataset Dataset}

\begin{figure*}[!t]
    \centering
    \includegraphics[width=1.0\linewidth]{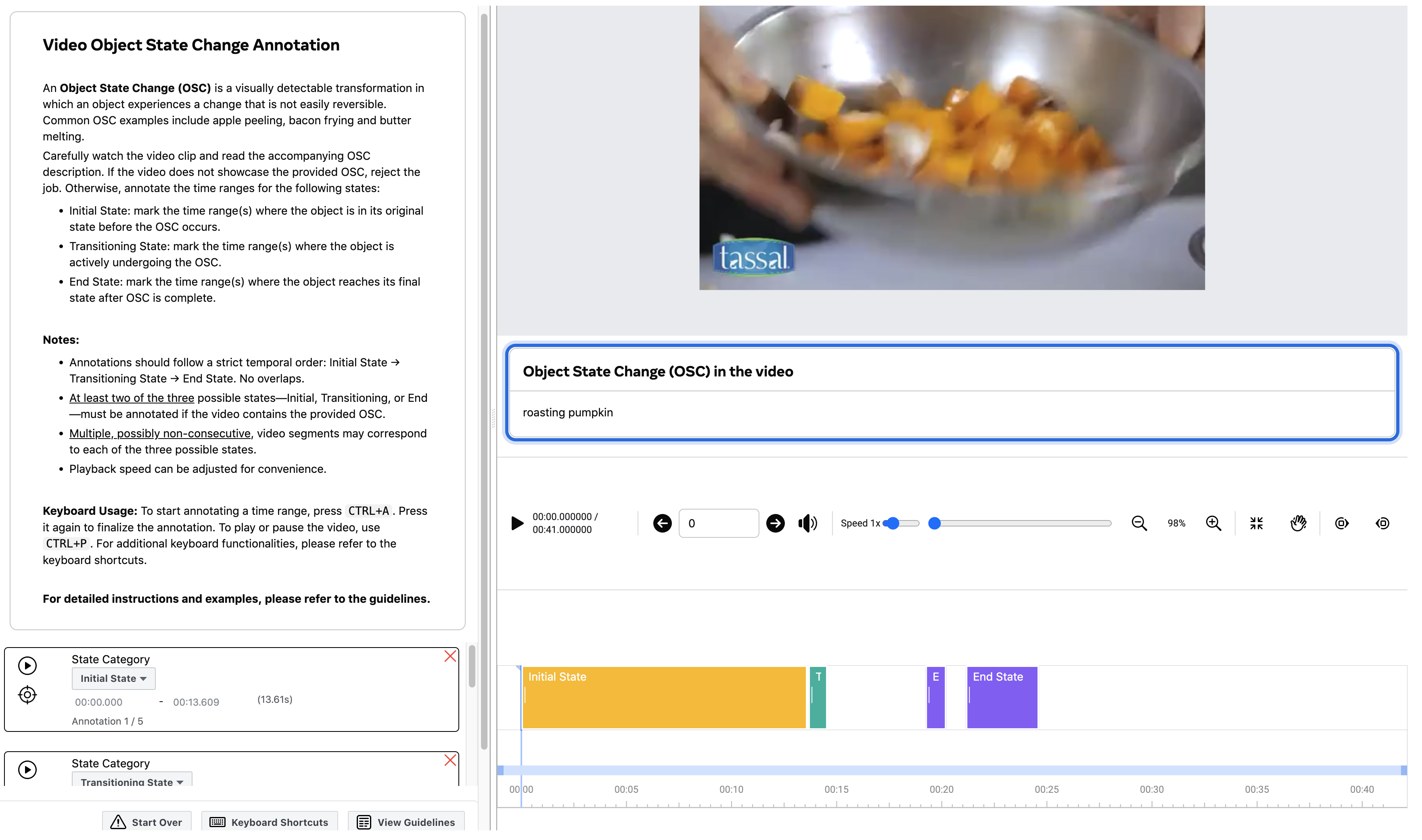}
    \caption{The annotation user interface. Annotators view a video paired with an OSC category, identified from the OSC mining process (outlined in Sec.~\ref{sec:pseudo_label}). They are instructed to either reject the video if it does not demonstrate the specified OSC, or to annotate the time ranges corresponding to the initial, transitioning, and end states of the OSC shown in the video.}
    \label{fig:annotation_ui}
\end{figure*}

\begin{table*}[!t]
\tablestyle{5pt}{1.1}
  \centering
\begin{tabular}{|l|p{10cm}|p{5cm}|}
\hline
State Transition & Objects (known) & Objects (novel) \\
\hline
blending & banana, egg, tomato, strawberry, garlic, butter, oat, sugar, milk, ice, onion, date, cashew, sauce, almond, cheese, mango & cream, pepper, avocado, ginger, coconut \\
\hline
browning & onion, chicken, garlic, meat, beef, sausage, butter, crust, bacon, pork, meatball, mushroom, tofu, turkey, steak, potato & bread, vegetable, sugar, apple, banana \\
\hline
chopping & onion, garlic, tomato, apple, parsley, carrot, pepper, mushroom, bacon, cilantro, spinach, cabbage, nut, banana, strawberry, cucumber, chocolate, rosemary, chive, shallot, peanut, vegetable, herb, kale, celery, mint, dill, mango, chicken, walnut, leaf, potato, jalapeno, zucchini, chili, egg, pecan, ginger, coriander, basil, avocado, broccoli, scallion, lettuce & cauliflower, almond, sausage, pineapple, date, leek, butter, chilies, capsicum, olive, thyme \\
\hline
coating & chicken, potato, fish, apple, tofu, bread, shrimp, cake & onion, butter, pasta \\
\hline
crushing & garlic, biscuit, oreo, cooky, potato, ice, walnut, ginger, strawberry, cracker, tomato, peanut, nut & almond, pepper, pineapple, onion \\
\hline
frying & onion, garlic, potato, bacon, chicken, tortilla, egg, fish, plantain, tofu, bread, mushroom, tomato, rice, sausage, batter, paneer, eggplant, shallot, beef, vegetable, shrimp, ginger, okra, pork, banana & cauliflower, carrot, steak, meat, pepper \\
\hline
grating & cheese, ginger, potato, carrot, garlic, nutmeg, orange, zucchini, cucumber, parmesan, chocolate, apple, lemon, onion & cauliflower, tomato, butter, mozzarella, coconut \\
\hline
grilling & chicken, steak, corn, fish, pineapple, salmon, onion, bread, tomato, zucchini, asparagus, bacon, eggplant & shrimp, sausage, cheese, potato, pepper \\
\hline
mashing & banana, potato, avocado, garlic, tomato, bean, butter, chickpea, strawberry & berry, egg, cauliflower \\
\hline
melting & butter, chocolate, cheese, sugar, marshmallow, ghee, caramel, jaggery, gelatin, margarine, mozzarella, shortening, candy & honey, cream, ice \\
\hline
mincing & garlic, ginger, onion, shallot, jalapeno, cilantro, parsley, beef, meat, scallion & pepper, tomato, carrot \\
\hline
peeling & banana, potato, apple, onion, garlic, plantain, egg, orange, ginger, carrot, cucumber, lemon, tomato, squash, avocado, mango, eggplant, pumpkin, shrimp, pear, beet, zucchini, prawn & pepper, shallot, peach, pineapple, kiwi \\
\hline
roasting & pepper, peanut, garlic, tomato, eggplant, potato, coconut, onion, nut, pumpkin, almond, chicken, vegetable, cauliflower, carrot, hazelnut, turkey, chickpea, corn, asparagus & broccoli, squash, beet, mushroom \\
\hline
rolling & dough, fondant, pastry, meatball, clay, cake, bread, pasta, tortilla & sausage, crust, cheese \\
\hline
sauteing & onion, garlic, mushroom, vegetable, carrot, ginger, celery, pepper, tomato, shallot & spinach, shrimp, chicken, potato \\
\hline
shredding & chicken, cheese, cabbage, carrot, potato, zucchini, beef, pork, meat, lettuce & coconut, mozzarella, parmesan, onion \\
\hline
slicing & onion, tomato, apple, potato, mushroom, garlic, lemon, banana, strawberry, cabbage, meat, zucchini, chicken, mango, pepper, cake, shallot, egg, sausage, watermelon, carrot, tofu, ginger, leek, beef, cucumber, scallion, eggplant, avocado, bread, pear, steak, pineapple, radish, peach, bacon & jalapeno, celery, butter, olive, mozzarella, orange, ham, lime, almond, cheese, pepperoni \\
\hline
squeezing & lemon, lime, orange, garlic, potato, spinach, avocado, zucchini, tomato, grapefruit, cabbage & ginger, cucumber, onion, tofu \\
\hline
whipping & cream, egg, butter, sugar, milk, potato, ganache, buttercream, batter, yogurt, frosting & mascarpone, strawberry, meringue \\
\hline
zesting & lemon, orange, lime, citrus, grapefruit & clementine \\
\hline
\end{tabular}
  \caption{The OSC taxonomy for \dataset\ encompasses 134 objects undergoing 20 distinct state transitions, resulting in 409 unique OSCs (318 known and 91 novel).}
  \label{tab:osc_taxonomy}
\end{table*}

\begin{figure*}[p]
    \centering
    \includegraphics[width=1.0\linewidth]{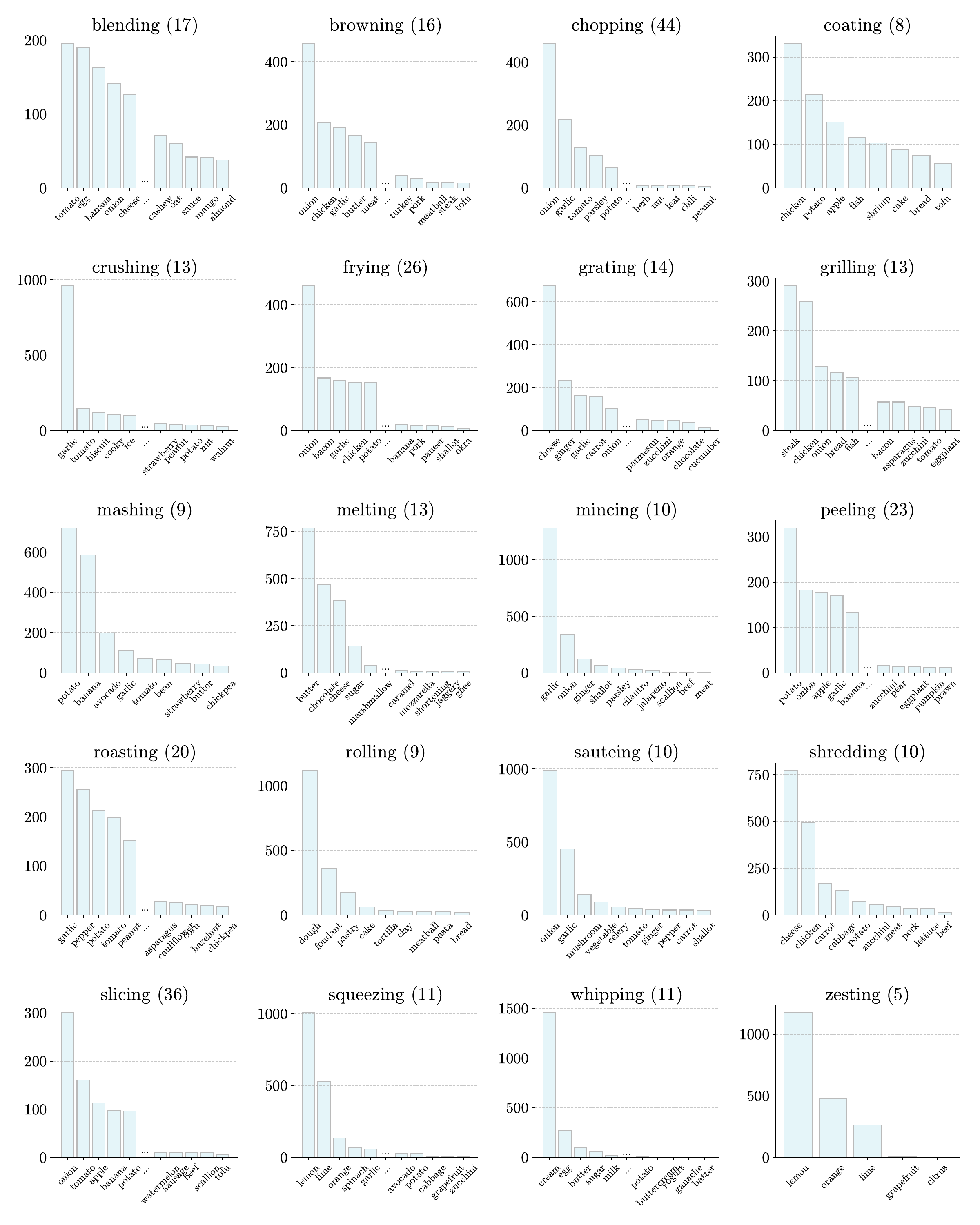}
    \caption{Data distribution of \dataset\ (Training). The y-axis denotes the number of annotated videos, and numbers in parentheses represent the count of unique objects associated with each state transition. Our data collection process mines video OSCs that authentically reflects the real-world's long-tail.}
    \label{fig:train_distribution_persc}
\end{figure*}

To establish the OSC taxonomy, we identify 20 most frequent state transitions and the objects associated with these state transitions that appear more than 200 times. Utilizing a 0.25 quantile threshold for each
state transition, we categorize the top 75\% frequent OSCs
as known and the bottom 25\% as novel, resulting in 318 known and 91 novel OSC categories in total. See Table \ref{tab:osc_taxonomy} for the complete OSC taxonomy. With these 318 known OSCs, we compose the training set of \dataset, encompassing 36,076 video segments from HowTo100M. Fig.~\ref{fig:train_distribution_persc} provides the detailed distribution of \dataset\ (Training).

\begin{figure*}[p]
    \centering
    \includegraphics[width=1.0\linewidth]{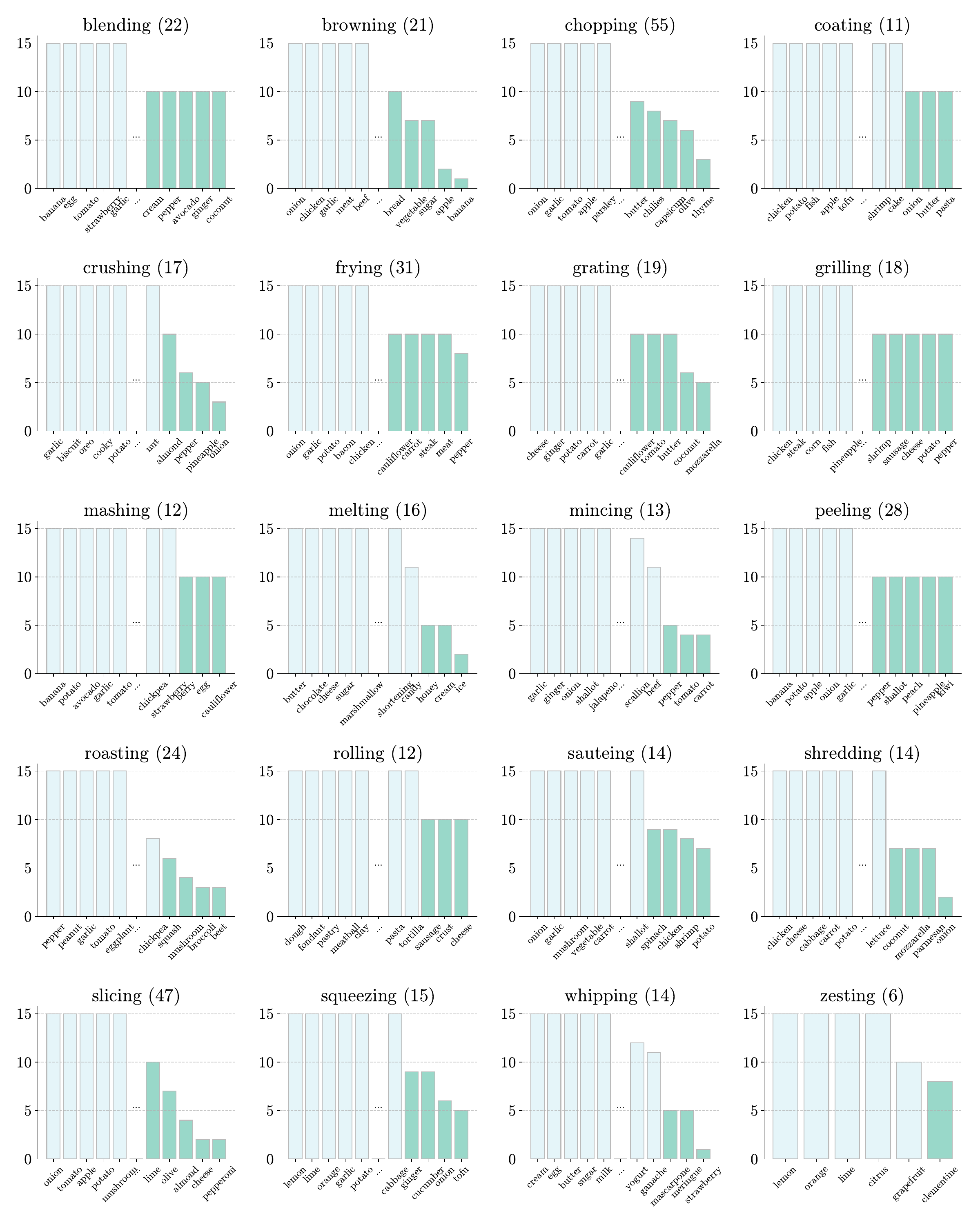}
    \caption{Data distribution of \dataset\ (Evaluation). Known and novel OSCs are shown in light blue and dark green, respectively. \dataset\ (Evaluation) presents a comprehensive evaluation benchmark, encompassing a diverse array of objects and state transitions.}
    \label{fig:ann_distribution_persc}
\end{figure*}

\begin{figure}[!thb]
    \centering
    \includegraphics[width=1.0\linewidth]{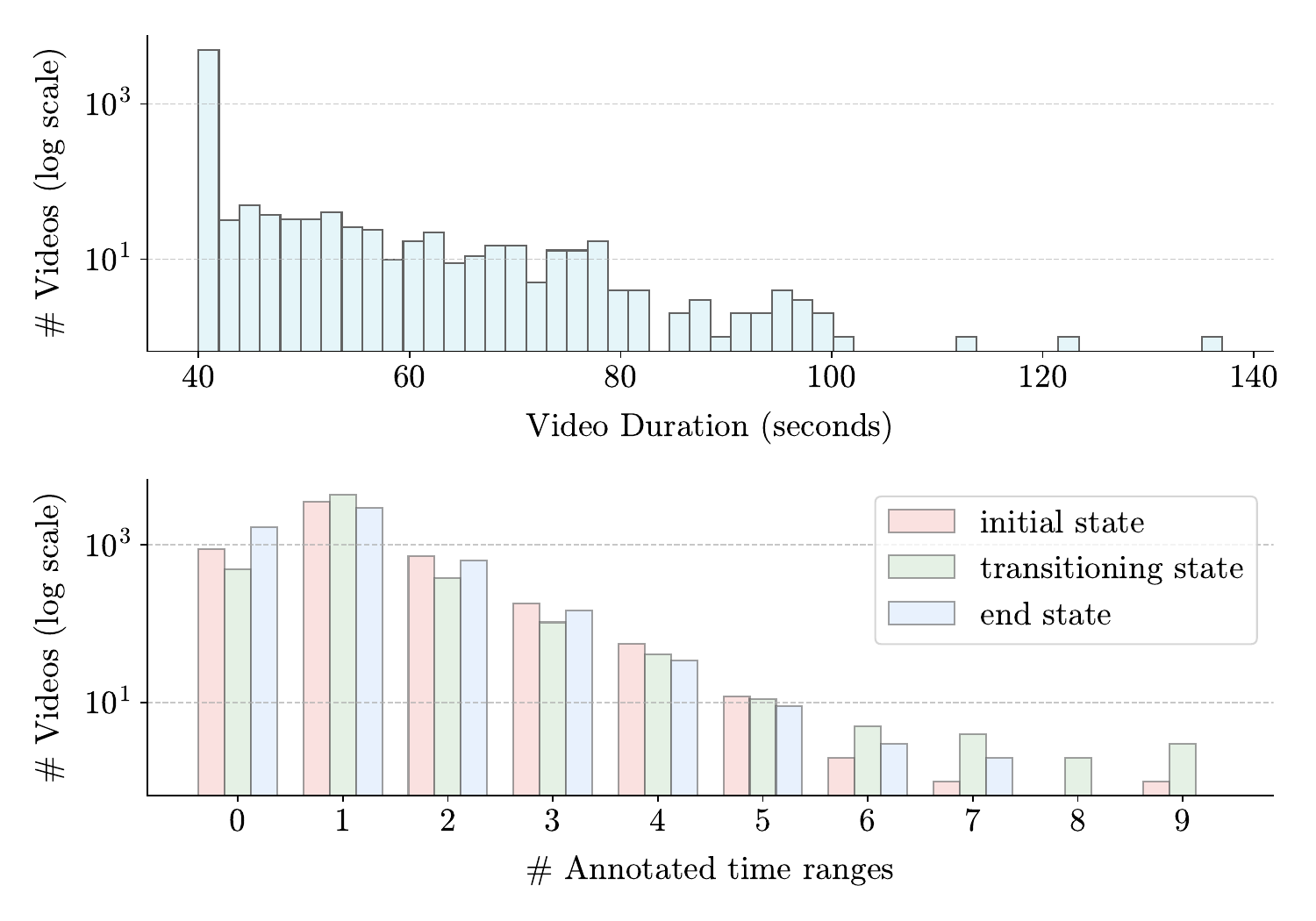}
    \caption{Distribution of video duration (upper) and number of annotated time ranges (lower).}
    \label{fig:duration_state_distribution}
\end{figure}

\subsection{Ground Truth Label Collection}
For evaluation, we collect annotations \KG{from 30 trained professional human annotators} for a subset of 5,423 video clips from \dataset, amounting to 62.5 hours of video. See Fig.~\ref{fig:annotation_ui} for the annotation user interface. We collect an average of 13.3 annotated videos per OSC category. The annotations for known and novel OSCs cap at 15 and 10, respectively. Fig.~\ref{fig:ann_distribution_persc} provides the detailed distribution of \dataset\ (Evaluation).

The breakdown of annotated time ranges within these videos is as follows: 19.8\% for the initial state, 25.9\% for the transitioning state, and 16.9\% for the end state, with the remaining categorized as non-OSC-related background. Fig.~\ref{fig:duration_state_distribution} shows the distribution of video duration (seconds) and the number of annotated time ranges per state in the \dataset\ (Evaluation) set. Importantly, the distribution demonstrates the granularity of our annotations, with a varied number of annotated time ranges per video. This is a result of our guidelines that instruct the annotators to exclude any time ranges where the OSC of interest is not observable, thereby ensuring precise labeling.

\section{Experiments}
\subsection{Experimental Setup}
\paragraph{Datasets} We conduct experiments on ChangeIt~\cite{souvcek2022look} and our proposed \dataset. \SX{Other related datasets are not included due to their small scale and differing OSC definition~\cite{alayrac2017joint}, unavailability for public access~\cite{liu2017jointly}, or the absence of OSC category / temporal labels necessary for our problem~\cite{grauman2022ego4d,saini2023chop}.}
Beyond the conventional split of ChangeIt, we introduce a novel split designed specifically for our open-world framework. The OSC taxonomy is detailed in Table~\ref{tab:changeit_openworld}. Note the inherent challenge of this setting: each state transition is associated with fewer than five objects. This scarcity of objects per state transition can significantly impede the model's ability to generalize the concept of states and state transitions without becoming overly dependent on specific objects, \KG{and reinforces the motivation for creating our new HowToChange dataset to augment this existing resource}.

\begin{table}[!th]
\tablestyle{5pt}{1.1}
  \centering
\begin{tabular}{|l|p{2.5cm}|p{2.5cm}|}
\hline
State Transition & Objects (known) & Objects (novel) \\
\hline
peeling & apple, dragon fruit, onion, pineapple & avocado, corn, eggs, garlic \\
\hline
frying & bacon & potatoes\\
\hline
pouring & beer, tea & juice, milk \\
\hline
wrapping & tortilla & gift/box \\
\hline
melting & butter & chocolate \\
\hline
cleaning & pan & shoes \\
\hline
tying & tie, ribbon & rope \\
\hline
cutting & tile & tree \\
\hline
\end{tabular}
  \caption{The OSC taxonomy for ChangeIt (open-world). Aligning with our open-world formulation, we propose a new split of ChangeIt~\cite{souvcek2022look} that randomly splits objects associated with the same state transition as known and novel. }
  \label{tab:changeit_openworld}
\end{table}

\paragraph{Evaluation}
For ChangeIt and ChangeIt (open-world), we adhere to the original dataset's evaluation protocol, reporting action and state precision@1 as the evaluation metrics. For our collected \dataset, while we also adopt state precision@1, due to our definition that positions the midpoint between the initial and end states as ``transitioning states'' rather than ``action'', our evaluation takes into account three distinct states, unlike the two states in ChangeIt. In addition, since precision@1 solely evaluates a single frame for each state within a video, we advocate for the use of F1 score and precision over all frames to ensure a more holistic evaluation. For each video, we compute the state precision@1, F1 score and precision for the present states (considering that a video might not always contain all three states: initial, transitioning, and end) and then compute their average over states. Subsequently, we average these values across videos within a state transition category and report the overall average for all state transitions. Lastly, for the two open-world datasets, we present these metrics on both known and novel OSCs.

\paragraph{Baselines}
% explain train and test difference
For results on ChangeIt, we reference the metrics as officially reported in their original papers. For results on ChangeIt (open-world), we reimplement the baselines to accommodate our newly introduced data split. As the LookForTheChange baseline~\cite{souvcek2022look} requires a model for every OSC category, when evaluating novel OSCs, we apply every model trained on OSCs with the same state transition and report best model performance. For the MultiTaskChange baseline~\cite{souvcek2022multi}, we train one multi-task model across all known OSCs and evaluate on both known and novel OSCs. On \dataset, %given that training 318 individual models (reflecting the 318 known OSC categories in \dataset) is impractical and we observe inferior results of these baselines under such conditions due to task competition,
\finalv{baseline results ~\cite{souvcek2022look, souvcek2022multi} were obtained using InternVideo features, not their originally proposed features, to ensure comparability.}
Aligning with our own approach, we adopt a shared vocabulary for both baselines. This means grouping OSCs with the same state transition as one category to enhance generalization. On all datasets, following their original papers, we enforce an additional casual ordering constraint during test time as we observe better performance of the baselines in this setting. \finalv{We adopt adaptive weights for baaselines on open-world ChangeIt using the values from their original papers but not on HowToChange due to no exemplar images.}
% Note that both baselines are trained on same data as ours, where no ground truth labels are available. While these baselines rely on a causal ordering constraint for supervisory signals, our \method\ innovatively employs VLMs to generate pseudo labels (Sec.~\ref{sec:pseudo_label}).
% For ours, we do not explicitly enforce this constraint and we observe our model already learns the ordering pattern during training (Fig. \ref{fig:viz_temporal_prediction}).

Regarding the three zero-shot baselines (i.e., CLIP~\cite{clip}, VideoClip~\cite{xu2021videoclip} and InternVideo~\cite{wang2022internvideo}), we adopt both the vision and language encoders to compute the similarity score between each (image/video, text) pair. The OSC state description text is same as adopted in our pseudo label generation process (Sec.~\ref{sec:pseudo_label}). Based on the similarity scores, we then conduct a grid search within each state transition to pinpoint the optimal threshold distinguishing background classes from OSC state categories, and report the best value.

\paragraph{Implementation}
For training pseudo label generation, we first employ GPT-4~\cite{openai2023gpt4} to automatically generate text descriptions for each OSC category in the dataset. We then assign pseudo labels to each video segment based on the similarity scores given by a CLIP~\cite{clip} model
% trained on LAION~\cite{schuhmann2022laion}
for ChangeIt and a VideoCLIP~\cite{xu2021videoclip} trained on HowTo100M~\cite{miech2019howto100m} for \dataset. We conduct a grid search for the two pseudo label thresholds $\delta$ and $\tau$ to identify the best value. We employ the AdamW optimizer with a learning rate of 1e-4 and a weight decay of 1e-4. Models are trained using a batch size of 64, over 50 epochs. Training takes a few hours on a NVIDIA A100.

To ensure a thorough evaluation, we train both single-task and multi-task variants of our approach.
% In the single-task version, we train an individual model for each state transition across the three datasets. For the multi-task variant, we train one model for all OSC classes in ChangeIt; one model for the 13 known OSC classes in ChangeIt (open-world), and one model for the 20 state transitions in \dataset.
We note that the approach is designed differently for ChangeIt and \dataset, due to their distinct characteristics: (1) For ChangeIt and ChangeIt (open-world), the term single-task denotes training a separate model for each OSC category (e.g. peeling apples), whereas multi-task denotes training a unified model for all OSC categories. Regarding baseline methods, LookforTheChange~\cite{souvcek2022look} aligns with the single-task paradigm while MultiTaskChange~\cite{souvcek2022multi} belongs to the multi-task one. (2) For \dataset, where each state transition is associated with a much broader range of objects, we adopt a shared state vocabulary to enhance model generalization, both for our approach and the baselines.
%(the benefit of a shared state vocabulary is demonstrated in the ablation study presented in Table \ref{tab:ablation}).
In this context, the single-task model is developed for each state transition (e.g, peeling) rather than each OSC (e.g. peeling apples). Consequently, a single-task model is already capable of identifying states for any OSCs that fall within the same state transition category. The multi-task model extends this concept to accommodate all 20 state transitions in \dataset. Both baselines, LookforTheChange~\cite{souvcek2022look} and MultiTaskChange~\cite{souvcek2022multi} fall into the single-task implementation as we observe worse performance and prohibitive long training time with the multi-task formulation.

Our multi-task model follows the same design as the single-task, with the modification of an expanded output label dimension to encapsulate all categories. Essentially, the multi-task variant can be conceptualized as a hierarchical classification problem. During testing, the model first determines the most probable state transition (e.g., peeling) based on prediction scores. Subsequently, it provides a prediction of fine-grained states for each time point (e.g., initial, transitioning and end state of peeling, or background).
It's important to note that while the single-task variant only performs the latter prediction step, the multi-task variant adds the ability to name the state transition. The multi-task model thus offers the benefit of a single, unified model that can predict OSC states for all categories of videos, eliminating the need for developing individual specialized models.

Finally, we emphasize that our model, irrespective of the variant, \emph{relies solely on video as input}. This is in contrast to VLM baselines \SX{(i.e., CLIP~\cite{clip}, VideoCLIP~\cite{xu2021videoclip} and InternVideo~\cite{wang2022internvideo})},
where the OSC text is required as input to calculate the cross-modality similarity.
% \KGnote{also write out their names here as they appear in the results table}

\subsection{Results} \label{sec:supp_result}

\begin{table}[!t]
\tablestyle{5pt}{1.1}
  \centering
  \begin{tabular}{lcccccc}
    \toprule
    \multirow{2}{*}{State Transition} & \multicolumn{2}{c}{F1 (\%)} & \multicolumn{2}{c}{Prec (\%)} & \multicolumn{2}{c}{Prec.@1 (\%)} \\
     & known & novel & known & novel & known & novel \\
    \midrule
   chopping & 46.5 & 44.1 & 43.7 & 42.4 & 58.3 & 58.2 \\
   slicing & 48.6 & 45.6 & 49.7 & 44.9 & 68.6 & 63.7 \\
   frying & 56.3 & 53.7 & 53.5 & 50.8 & 61.2 & 54.5 \\
   peeling & 49.0 & 42.4 & 51.4 & 45.8 & 65.6 & 57.7 \\
   blending & 42.2 & 45.2 & 43.4 & 50.7 & 59.1 & 66.7 \\
   roasting & 36.5 & 40.8 & 40.3 & 44.4 & 59.2 & 64.6 \\
   browning & 44.9 & 51.5 & 46.3 & 54.4 & 55.1 & 60.5 \\
   grating & 52.5 & 51.3 & 51.6 & 50.4 & 66.6 & 65.0 \\
   grilling & 54.6 & 53.7 & 54.0 & 49.6 & 67.8 & 61.0 \\
   crushing & 39.2 & 32.2 & 38.3 & 28.0 & 58.9 & 52.8 \\
   melting & 34.9 & 36.8 & 35.1 & 38.2 & 46.6 & 38.9 \\
   squeezing & 54.4 & 54.6 & 54.0 & 54.6 & 61.0 & 66.1 \\
   sauteing & 47.7 & 36.4 & 47.1 & 41.4 & 56.3 & 46.0 \\
   shredding & 53.3 & 41.8 & 52.8 & 44.8 & 66.6 & 58.7 \\
   whipping & 45.1 & 43.1 & 46.9 & 44.3 & 57.8 & 45.5 \\
   rolling & 39.7 & 32.6 & 43.5 & 39.3 & 62.2 & 60.0 \\
   mashing & 52.4 & 52.0 & 52.2 & 53.8 & 66.7 & 69.4 \\
   mincing & 45.3 & 37.2 & 41.7 & 32.1 & 54.7 & 55.1 \\
   coating & 35.5 & 30.0 & 37.8 & 28.5 & 55.1 & 48.3 \\
   zesting & 49.6 & 36.2 & 49.3 & 35.3 & 65.9 & 70.8 \\
    \midrule
   Average & 46.4 & 43.1 & 46.6 & 43.7 & 60.7 & 58.2 \\
    \bottomrule
  \end{tabular}
  \caption{Detailed per-state-transition results of \method\ on \dataset.}
  \label{tab:result_detailed}
\end{table}

\begin{table*}[!t]
\tablestyle{4pt}{1.1}
  \centering
\begin{tabular}{lcccccccccccc}
    \toprule
     \multirow{3}{*}{Method} & \multicolumn{2}{c}{ChangeIt} &  \multicolumn{4}{c}{ChangeIt (open-world)} & \multicolumn{6}{c}{\dataset} \\
    & \multirow{2}{*}{\makecell[c]{State \\Prec.@1}} & \multirow{2}{*}{\makecell[c]{Action \\Prec.@1}}& \multicolumn{2}{c}{State Prec.@1} & \multicolumn{2}{c}{Action Prec.@1} & \multicolumn{2}{c}{F1 (\%)} & \multicolumn{2}{c}{Prec (\%)} & \multicolumn{2}{c}{Prec.@1 (\%)} \\
    &  &  & known & novel & known & novel & known & novel & known & novel & known & novel \\
    \midrule
\method\ (multi-task) & 0.44 & 0.69 & 0.43 & 0.29 & 0.75 & 0.63 &  40.7 & 37.9 & 41.8 & 39.0 & 56.8 & 54.8 \\
\method\ (single-task) & 0.57 & 0.84 & 0.56 & 0.48 & 0.89 & 0.82 & 46.4 & 43.1 & 46.6 & 43.7 & 60.7 & 58.2\\
    \bottomrule
  \end{tabular}
  \caption{A comparison of the single-task and multi-task variant of \method. The multi-task variant offers the benefit of a single, unified model capable of predicting fine-grained OSC states for videos of all OSC categories, while the single-task variant is optimized for each individual OSC and demonstrates superior performance. }
  \label{tab:result_multitask}
\end{table*}

\begin{figure*}[!t]
    \centering
    \includegraphics[width=1.0\linewidth]{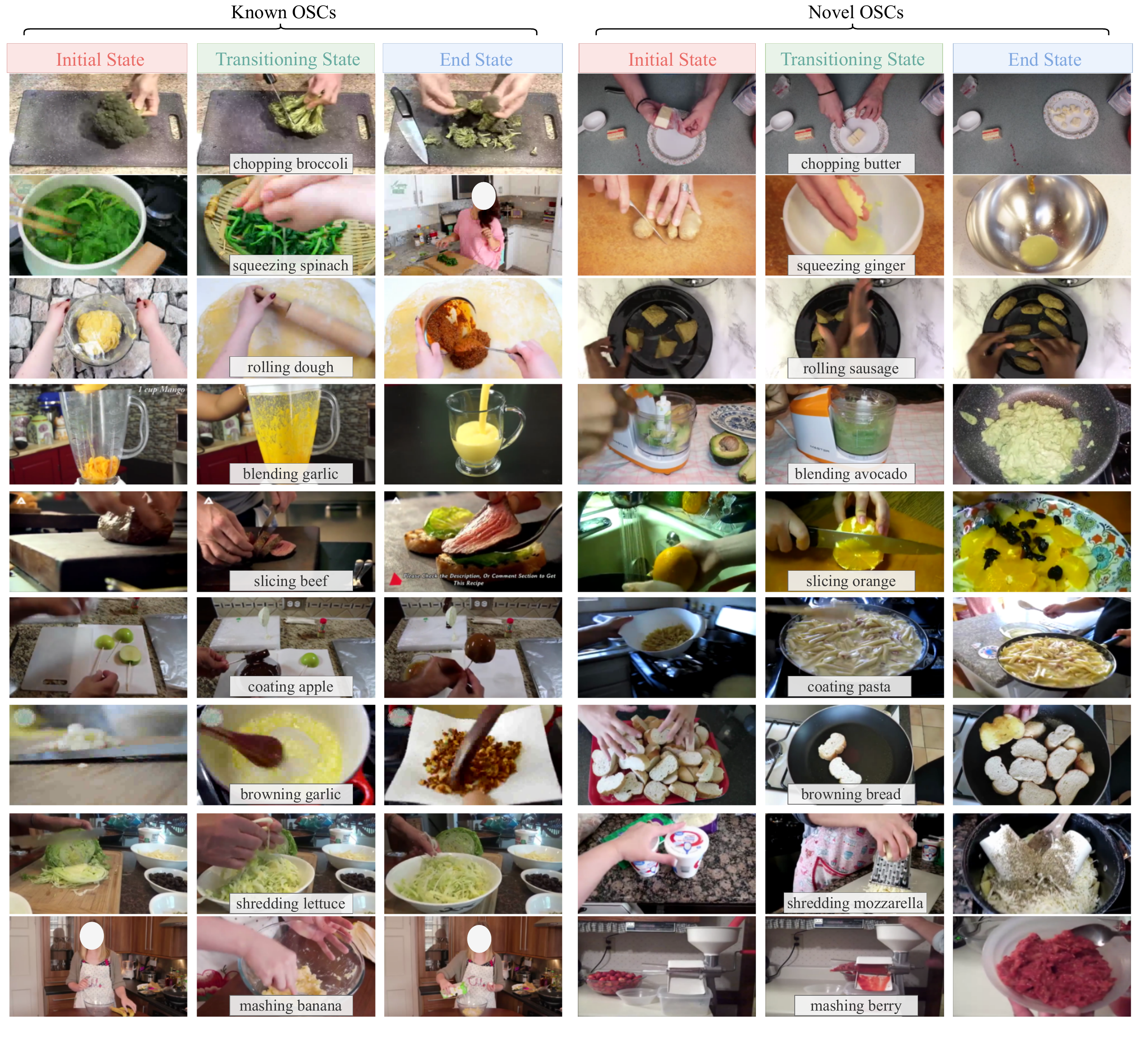}
    \vspace*{-10mm}
    \caption{Top-1 frame predictions given by \method\ for the initial, transitioning, and end states, on \dataset (Evaluation). \method\ not only accurately localizes the three fine-grained states for known OSCs, but also generalizes this understanding to novel objects, which are not observed during training. \KGnote{should we add a figure of failure cases and briefly explain them?}}
    \label{fig:viz_pred2}
\end{figure*}

\paragraph{Detailed per-state-transition results} Supplementing Table~\ref{tab:combined_results} in the main paper, we provide a detailed breakdown of \method's performance on \dataset\ per state transition in Table~\ref{tab:result_detailed}. We observe superior results in transitions like mashing, squeezing, and grilling, while transitions such as melting and coating show comparatively weaker performance, possibly due to the ambiguity in their OSC states. In addition, the known-novel OSC gap is smallest for chopping, grating and mashing, whereas shredding and sauteing exhibit larger performance discrepancies. \SX{Intuitively, state transitions like chopping and mashing share more invariant representations across objects, with objects consistently going from whole to pieces or a mashed state. In contrast, shredding and sauteing may demonstrate less consistent transformation patterns across different objects, leading to greater variability and thus larger performance discrepancies.}
% \KG{Intuitively, objects that look particularly different when.... [add a sentence here to point out why this larger discrepancy for such verbs makes sense...]}
We hope these results provide insight for further analysis and development in this area.

\begin{figure}[!t]
    \centering
    \includegraphics[width=1.0\linewidth]{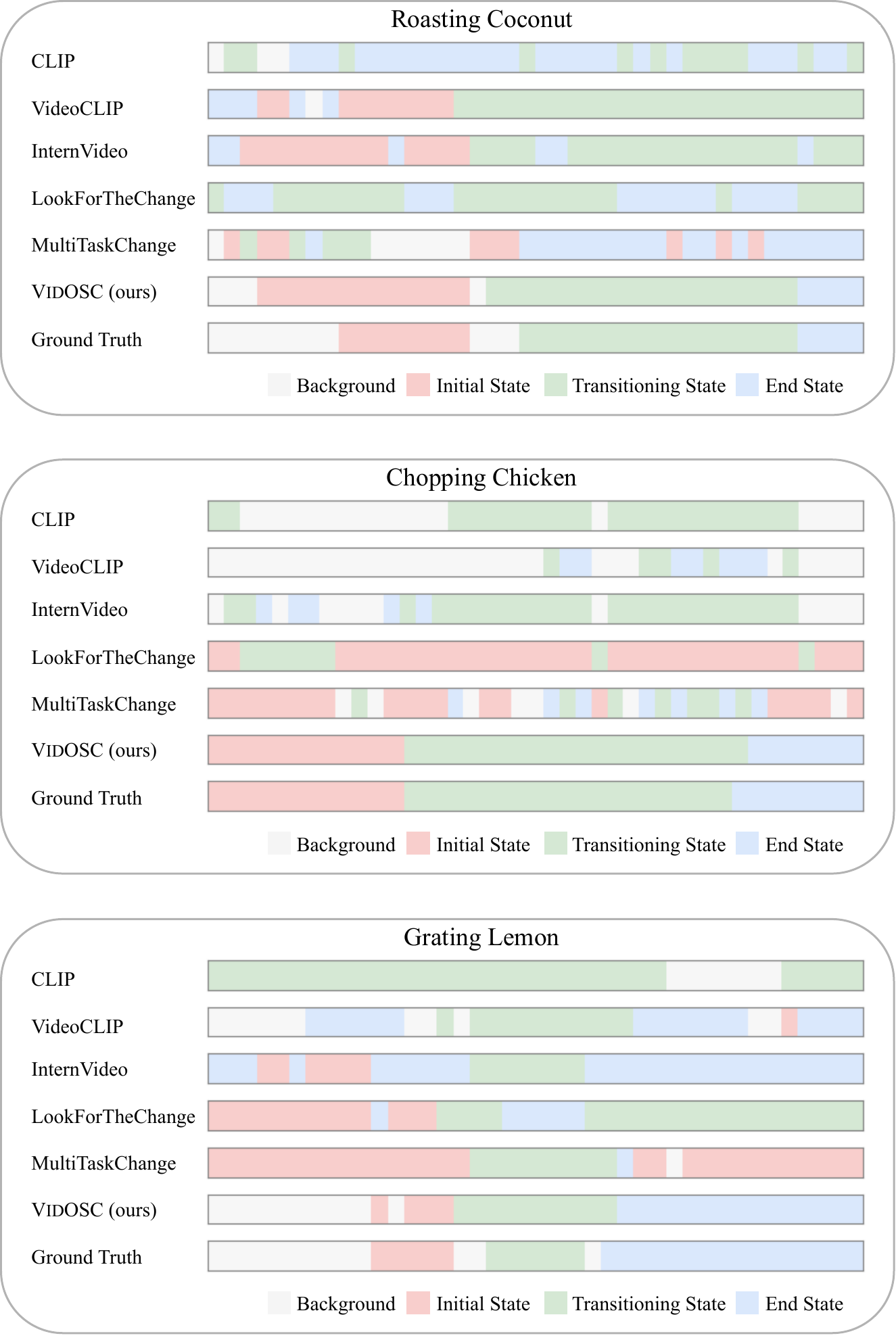}
    \caption{Comparison of model predictions on \dataset\ (Evaluation). The x-axis represents temporal progression through the video. \method\ gives temporally smooth and coherent predictions that best align with the ground truth, significantly outperforming baselines in capturing the video's global temporal context.}
    \label{fig:viz_temporal2}
\end{figure}

\paragraph{Single-task vs Multi-task} %\SX{As noted in Ln 459 of the main paper, we train both single-task and multi-task variants of \method.} We report performance of \method\ (single-task) in the main paper, and provide its comparison with the multi-task variant in Table \ref{tab:result_multitask}.
We train both single-task and multi-task variants of \method, and compare their performance in Table \ref{tab:result_multitask}.
While the multi-task model offers the convenience of a unified framework that can handle various state transitions simultaneously, they generally underperform their single-task counterparts. This performance disparity is a long-standing problem in multi-task learning and could stem from multiple factors such as varied convergence rates and potential competition among different OSCs, suggesting promising areas for future research. In terms of the comparison of \method\ (multi-task) with the MultiTaskChange baseline~\cite{souvcek2022multi}, for ChangeIt (open-world), \method\ achieves a +0.02 increase in state precision@1 and +0.03 in action precision@1 for known OSCs, and a +0.07 and +0.01 increase for novel OSCs, respectively. For the standard ChangeIt dataset, our reimplementation of MultiTaskChange based on their officially released code achieves a state precision@1 of 0.40 and an action precision@1 of 0.69, lower than the original paper's reported 0.49 and 0.80. \method\ (multi-task) surpasses the reproduced numbers with a 0.04 improvement in state precision@1.
% In conclusion, across both known and novel OSCs, single-task and multi-task paradigms, \method\ consistently outperform all baselines.
\SX{Lastly, we note that \dataset\ features closely related state transitions (such as crushing and mashing, melting and browning) as well as fine-grained variations within a general transition, (such as various cutting ways: chopping, slicing and mincing). These variations present both challenges and opportunities for the advancement of multi-task models, particularly in modeling the similarities and fine distinctions among different state transition categories. We leave it as future work.}
\KGnote{are we referrign only to the reproduced model, not their published numbers?  not sure which table these comments go with because in table 7 we have only VidOSC?} \SXnote{only the reproduced model. deleted that sentence about results.}

% We highlight the trade-off between the single-task and multi-task variants of \method. The multi-task model offers the convenience of a unified framework that can handle various state transitions simultaneously, thus simplifying test-time requirements. Unlike their single-task counterparts, which are designed to evaluate one specific OSC, the multi-task model allows for the direct input of videos without necessitating specification of the state transition category. It autonomously infers the most probable state transition category (e.g., melting) and subsequently predicts the fine-grained OSC state (e.g., initial / transitioning / end state of melting). Despite their versatility, multi-task models tend to underperform in generating optimal predictions for individual tasks, as observed both in \method\ and the comparison between MultiTaskChange and LookForTheChange on ChangeIt (open-world). This disparity might be attributed to varying convergence rates and potential task competition among different OSCs, an inherent challenge in multi-task learning~\cite{xue2023egocentric}. In conclusion, across both known and novel OSCs, single-task and multi-task paradigms, \method\ consistently outperform all baselines.

\paragraph{Further Qualitative Results}
% \SX{As noted in Ln 516 in the main paper,
We provide more qualitative results supplementing Fig.~\ref{fig:viz_prediction} and Fig.~\ref{fig:viz_temporal_prediction} in the main paper. Fig.~\ref{fig:viz_pred2} showcases more examples of \method's top-1 predictions on \dataset\ (Evaluation). \method\ gives correct predictions for various state transitions, across both known and novel objects. In addition, Fig.~\ref{fig:viz_temporal2} provides more examples of \method's predictions from a global perspective. Compared with all approaches, \method\ consistently delivers temporally coherent predictions that closely align with the ground truth labels. All these results help demonstrate the strong performance of \method.

\begin{figure}[!t]
    \centering
    \includegraphics[width=1.0\linewidth]{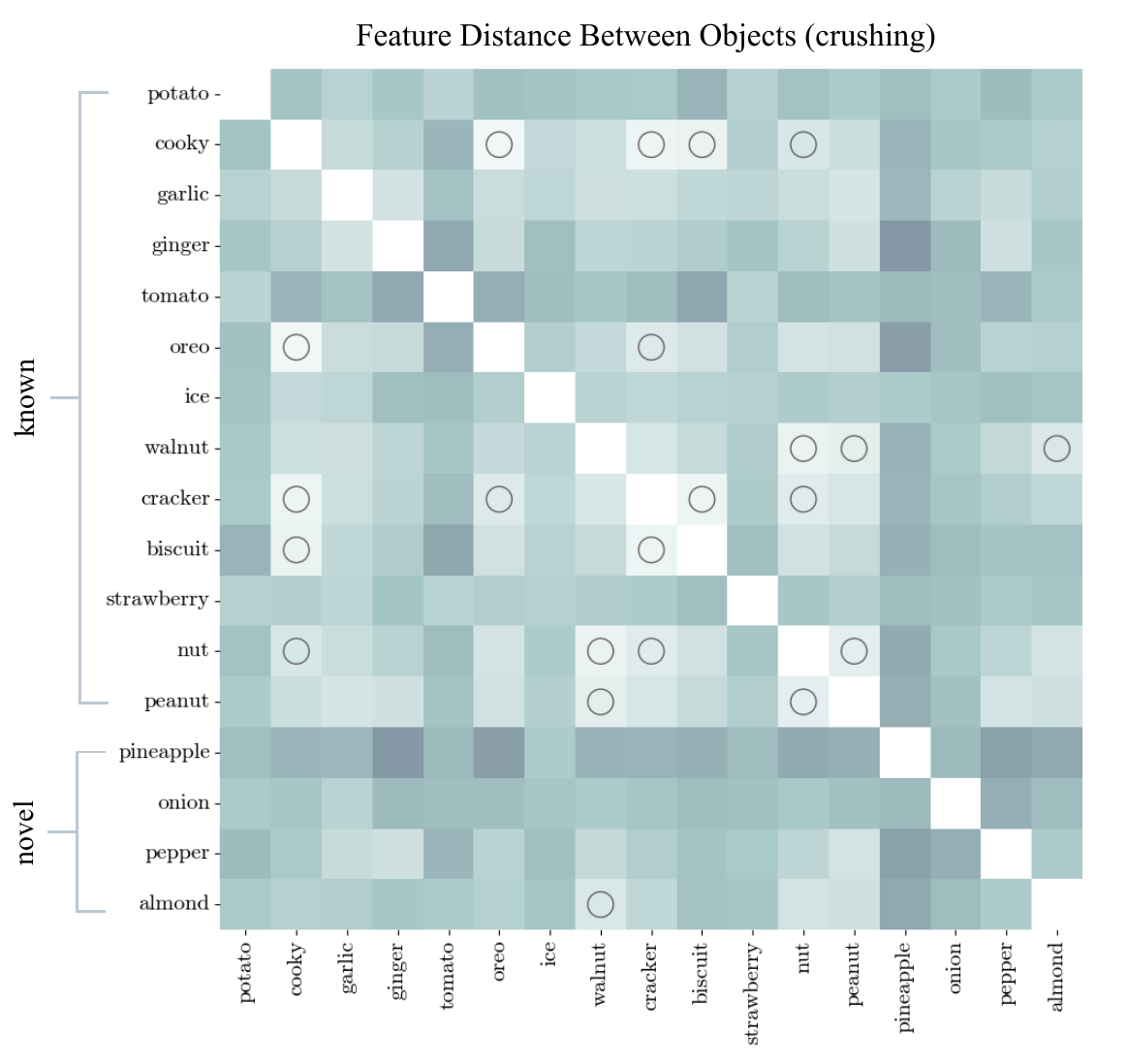}
    \caption{Distance matrix between object features produced by \method\ during the ``crushing'' process. A lighter color indicates smaller feature distance, and object pairs with relative small distance are marked by a circle. The heatmap offers interpretability on object relations during a state transition and provides insight on how the model generalizes from known to novel objects. }
    \label{fig:viz_objrelation}
\end{figure}

\paragraph{Interpretability on Object Relations}
% \SX{Elaborating on Ln 517 of the main paper,}
\method\ provides interpretable insights on how object relate to each other during specific state transitions. To illustrate this, we calculate features that belong to the transitioning state of ``crushing'', averaged over objects across all test videos. We then compute a feature distance matrix from these object features, as depicted in Fig.~\ref{fig:viz_objrelation}. While \method\ is purely video-based and has no access to ground truth object names, the feature embeddings it produces well captures the relations between each object pair during the ``crushing'' transition. For instance, crushing cracker is more similar to crushing oreo or biscuit than to crushing pineapple, which aligns with our intuition. Furthermore, the heatmap reveals how \method\ effectively leverages known object relationships to reason about novel objects. For example, the novel object ``almond'' is closet in feature space to ``walnut'' and ``peanut'' among all known objects.

\begin{table}[!t]
\tablestyle{2pt}{1.1}
  \centering
\begin{tabular}{lcccccc}
    \toprule
    \multirow{2}{*}{Method} & \multicolumn{2}{c}{F1 (\%)} & \multicolumn{2}{c}{Prec (\%)} & \multicolumn{2}{c}{Prec.@1 (\%)} \\
     & known & novel & known & novel & known & novel \\
    \midrule
    CLIP~\cite{clip} & 26.9 & 25.4 & 27.3 & 26.6 & 47.5 & 47.5\\
    \method\ (CLIP) & 35.5 & 34.1 & 38.6 & 36.3 & 51.1 & 48.5\\
    \hdashline
    Improvement & +8.6 & +8.7 & +11.3 & +9.7 & +3.6 & +1.0\\
    \midrule
    VideoCLIP~\cite{xu2021videoclip} & 36.6 & 34.3 & 39.7 & 38.5 & 48.3 & 44.8\\
    \method\ (VideoCLIP) & 46.4 & 43.1 & 46.6 & 43.7 & 60.7 & 58.2 \\
    \hdashline
    Improvement & +9.8 & +8.8 & +6.9 & +5.2 & +12.4 & +13.4\\
    \bottomrule
  \end{tabular}
  \caption{A comparison of \method\ using pseudo labels provided by CLIP~\cite{clip} and VideoCLIP~\cite{xu2021videoclip}. \SX{\method\ is a flexible framework can be combined with different VLMs. Employing a better VLM (VideoCLIP over CLIP) further enhances \method's performance.}
  % \KGnote{add a punchline here, like this shows that the InternVideo features used in the main paper results are the strongest....?}
  }
  \label{tab:result_clip}
\end{table}

\begin{figure}[!t]
    \centering
    \includegraphics[width=1.0\linewidth]{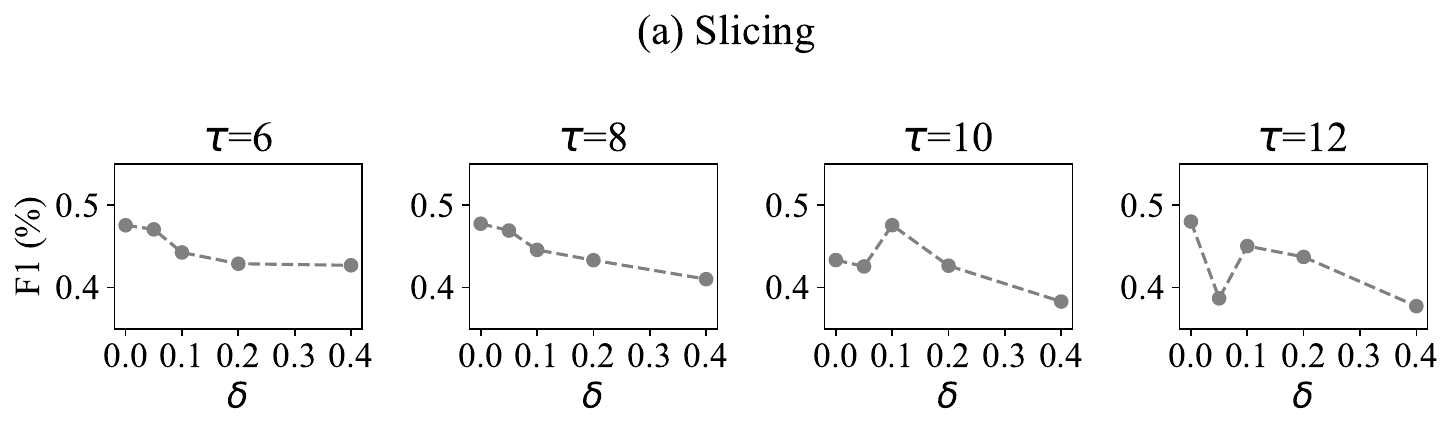}
    \includegraphics[width=1.0\linewidth]{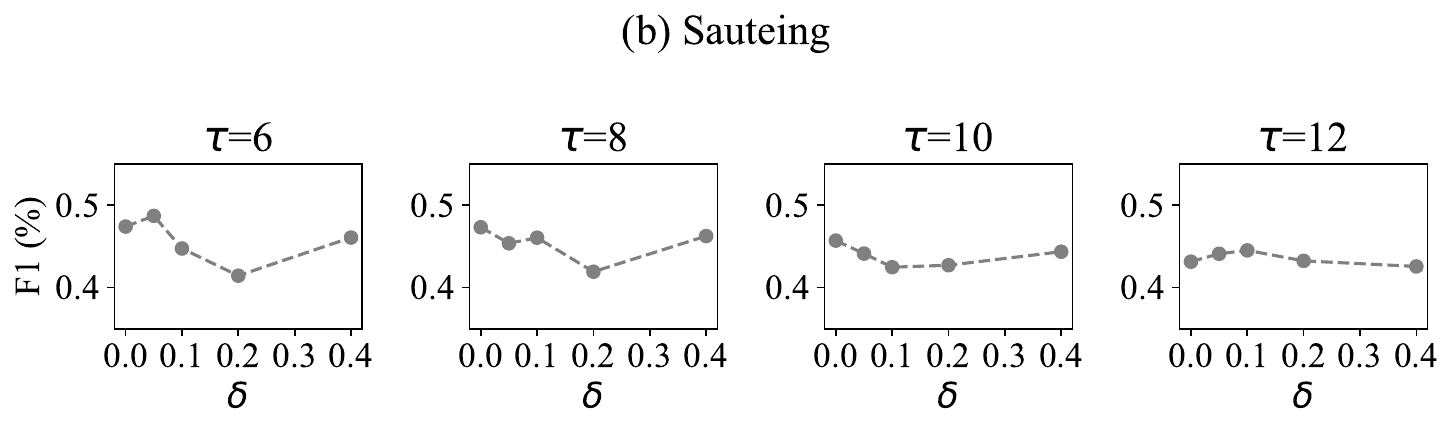}
    \caption{Analysis of pseudo label thresholds $\delta$ and $\tau$ for the state transition of (a) slicing and (b) sauteing. }
    \label{fig:pl_threshold}
\end{figure}

\paragraph{Pseudo Label Analysis} %\SX{As noted in Ln 528-529 of the main paper,}
\finalv{The pseudo label thresholds $\delta$ and $\tau$ in Section \ref{sec:pseudo_label} are decided via a hyperparameter search. Figure \ref{fig:pl_threshold} illustrates the process, showing F1 scores for different $\delta$ and $\tau$ for the state transition of slicing and sauteing. According to this analysis, we set pseudo label thresholds $\tau$ = 12 and $\delta$ = 0 for the state transition of slicing and $\tau$ = 6 and $\delta$ = 0.05 for sauteing. We repeat this process for all other state transitions and for zero-shot baselines as well for a fair comparison.}

\begin{table}[!t]
\tablestyle{2pt}{1.1}
  \centering
\begin{tabular}{lcccccc}
    \toprule
    \multirow{2}{*}{Method} & \multicolumn{2}{c}{F1 (\%)} & \multicolumn{2}{c}{Prec (\%)} & \multicolumn{2}{c}{Prec.@1 (\%)} \\
     & known & novel & known & novel & known & novel \\
    \midrule
 \method\ (no ordering) & 41.1 & 36.6 & 41.7 & 37.4 & 52.1 & 47.5 \\
    \method & 46.4 & 43.1 & 46.6 & 43.7 & 60.7 & 58.2 \\
    \bottomrule
  \end{tabular}
  \caption{A comparison of \method\ with and without ordering constraint enforced in pseudo label generation. Enforcing causal ordering leads to better pseudo labels and performance gains.}
  \label{tab:pl_ordering}
\end{table}

\finalv{In addition, we experiment with no ordering constraint enforced in pseudo label generation (Section \ref{sec:pseudo_label}) on HowToChange. Table \ref{tab:pl_ordering} underscores the positive impact of enforcing causal ordering, since otherwise the VLM-derived labels would be noisier and unordered in nature.}

Lastly, we compare the performance of \method\ using pseudo labels generated by two different VLMs, CLIP~\cite{clip} and VideoCLIP~\cite{xu2021videoclip} on \dataset. As demonstrated in Table~\ref{tab:result_clip}, \method\ learns to generalize and improve upon the pseudo labels it receives during training, outperforming the VLM baseline by a great margin. Notably, VideoCLIP yields better performance than CLIP. Correspondingly, \method\ incorporating VideoCLIP also surpasses \method\ using CLIP, achieving additional gains over the VideoCLIP baseline. This underscores the potential of \method: it can be synergistically combined with any advanced VLM to further augment performance.

\finalv{\paragraph{Task-specific model vs general VLMs.} We conclude with a discussion comparing \method\ with all-purpose VLMs.

We highlight that our \method\ addresses unique challenges in the open-world video OSC problem, which current VLMs are not yet equipped to handle. \emph{Long video temporal reasoning.} Our task involves understanding long videos (input videos range from 40 to 140 seconds, as shown in Figure \ref{fig:duration_state_distribution}). This requires long temporal reasoning beyond the capabilities of current VLMs, which are primarily image-based or limited to processing short video clips. For instance, the state-of-the-art video foundation model InternVideo~\cite{wang2022internvideo}, which we use for feature extraction and as a baseline, is constrained to processing short clips of a few seconds. \emph{Fine-grained state understanding.} The core of our challenge lies in distinguishing fine-grained states within an OSC process. This level of detail requires a nuanced understanding that general VLMs currently lack. While they may excel in recognizing objects and basic actions, discerning subtle state changes in a process is a frontier yet to be fully explored by these models.

To substantiate our claims, we experiment with the advanced GPT-4V~\cite{openai2023gpt4}. When tasked with predicting OSC states for a 40-second video, GPT-4V fails to produce meaningful outputs, as shown by replies like ``I'm sorry, but I can't provide assistance with the task as described.'', ``I cannot process the request as it involved 40 separate images.'' or ``Unfortunately, I cannot assist with labeling or categorizing images in sequences.'' This underscores the current limitations of VLMs in long video modeling. Simplifying the task to single-frame state classification, we present prediction results of GPT-4V (on 3 individual runs) and ours in Figure \ref{fig:viz_temp_gpt4v}. While GPT-4V correctly classifies the background category in most cases, it shows great instability in distinguishing the three OSC states. This highlights its limitations in fine-grained state understanding. Note that we do not have access to GPT-4V's underlying features and are limited to interacting via API.

\begin{figure}[!t]
    \centering
    \includegraphics[width=0.9\linewidth]{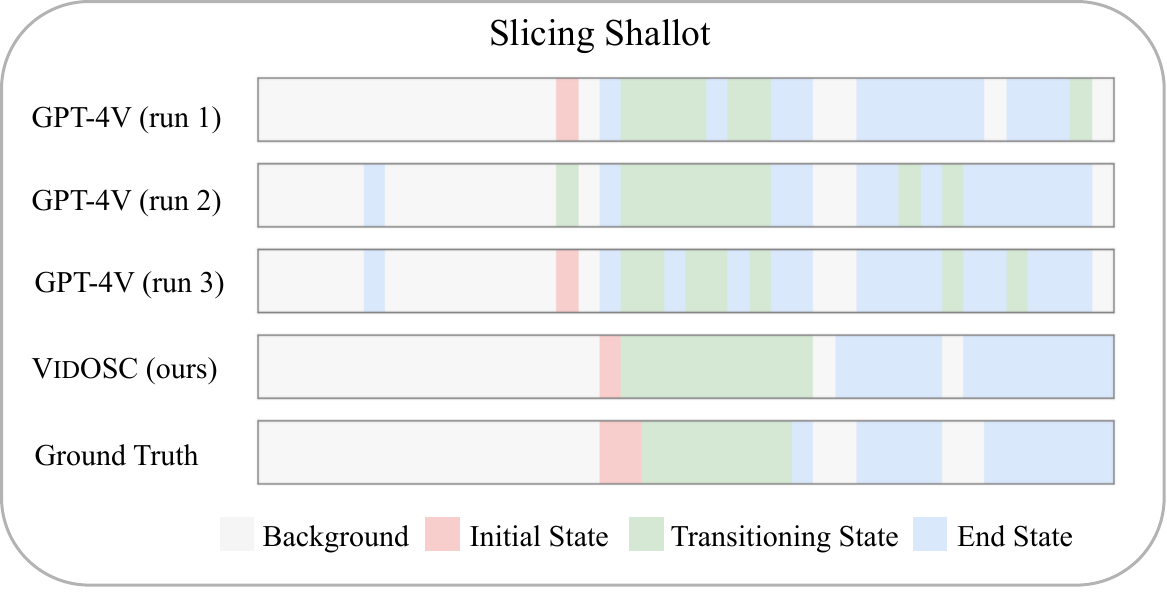}
    \caption{Comparison of \method\ with GPT4-V on a 40-second test video. \method\ provides more temporally coherent predictions.}
    \label{fig:viz_temp_gpt4v}
\end{figure}

Looking forward, we fully acknowledge and embrace the power of VLMs, which drives our automatic pseudo labeling approach. Benefiting from the rapid progress of general-purpose VLMs, \method\ is specialized in long and fine-grained video understanding, an area uncharted by VLMs; our work helps lay exactly the missing groundwork.}

% WARNING: do not forget to delete the supplementary pages from your submission 
% \input{sec/X_suppl}
\end{document}